\documentclass[journal]{IEEEtran}
\usepackage[T1]{fontenc}
\usepackage[utf8]{inputenc}
\usepackage{amsmath,amssymb,graphicx,booktabs,array,longtable,calc,seqsplit,xurl,needspace}
\usepackage[hidelinks]{hyperref}
\providecommand{\tightlist}{\setlength{\itemsep}{0pt}\setlength{\parskip}{0pt}}
\providecommand{\pandocbounded}[1]{#1}
\title{GeoPilot: Diagnosis-Guided Interventions and Geometric Evaluation for Aerial Reconstruction}
\author{Pengyu~Yin, Jie~Pan, Junxiang~Tan, Mengqi~Fu, Kun~Cheng, and~Yifan~Liu%
\thanks{This work was supported by the Aerospace Information Research Institute, Chinese Academy of Sciences, under Grant E6E3190301.}%
\thanks{\textit{(Corresponding author: Junxiang Tan.)}}%
\thanks{Pengyu Yin, Jie Pan, Junxiang Tan, and Mengqi Fu are with the Aerospace Information Research Institute, Chinese Academy of Sciences, Beijing 100094, China (e-mail: yinpy@aircas.ac.cn; panjie@aircas.ac.cn; tanjx@aircas.ac.cn; 1043879125@qq.com).}%
\thanks{Kun Cheng and Yifan Liu are with the School of Electronic Engineering, Beijing University of Posts and Telecommunications, Beijing 100876, China (e-mail: kcheng@bupt.edu.cn; yifanliu@bupt.edu.cn).}}
\hypersetup{pdfauthor={Pengyu Yin, Jie Pan, Junxiang Tan, Mengqi Fu, Kun Cheng, Yifan Liu}}
\begin{document}
\raggedbottom
\maketitle
\begin{abstract}
When a mesh reconstructed from aerial remote-sensing imagery disagrees with reference geometry, deciding what to change requires a diagnosis of the reconstruction process. We present GeoPilot, a workflow that turns diagnostics and prior observations into executable program or camera-configuration interventions. A large language model proposes the change, numerical tools reconstruct the mesh, and an external evaluator measures its geometry under a stated coordinate and reference protocol. The main case tests a shared-intrinsics hypothesis proposed before the comparison. On 224 Metashape-raster development images, shared intrinsics improved all six native-coordinate measures and reduced mean surface-to-reference distance by 7.04\%. Reference-assisted alignment fitted and evaluated on separate within-scene checkerboard samples reversed the mean-distance ranking, limiting that gain to native-frame delivery. A separate matched comparison on publisher-undistorted images favored fixed published cameras on all six measures, reducing mean distance by 12.11\%; the cameras retain upstream image and laser-scanning adjustment information. With one reconstruction per configuration, these comparisons provide conditional evidence for camera handling. Three end-to-end runs each delivered a mesh and six-metric evaluation from all 327 images of a second development scene.
\end{abstract}
\section{Introduction}\label{introduction}

Geometric meshes reconstructed from aerial remote-sensing imagery depend on camera calibration, image preparation and surface reconstruction. A reconstruction may complete successfully while its mesh still disagrees with reference geometry. The next change could concern meshing, calibration or image preparation, depending on the suspected cause. COLMAP and OpenMVS provide the numerical components \cite{ref1,ref2}; choosing which component to change requires interpreting the diagnostics and testing the resulting mesh.

The shared-intrinsics case studied here illustrates that decision on the Metashape raster track of Dataset-1, denoted M. The reconstruction assigned a separate intrinsic model to each image of a homogeneous image set. Camera-center residuals directed attention to calibration flexibility, and the agent proposed sharing intrinsics before the matched comparison. The shared-intrinsics reconstruction improved all six native-coordinate measures, yet reference-assisted alignment reversed the mean-distance ranking. A useful intervention therefore needs both a reason for trying it and a clear account of what its geometric result supports.

We present GeoPilot, a workflow that connects these steps. A large language model (LLM) reads diagnostics, tool descriptions and selected prior results, then proposes a checked numerical program or a versioned camera configuration. Numerical tools execute the proposal, and an external evaluator measures the delivered mesh under a stated coordinate and reference protocol. Experience entries preserve the change, its outcome and the conditions under which it was observed (Figure 1).

\begin{figure*}[!t]
\centering
\pandocbounded{\includegraphics[width=\linewidth,height=0.62\textheight,keepaspectratio]{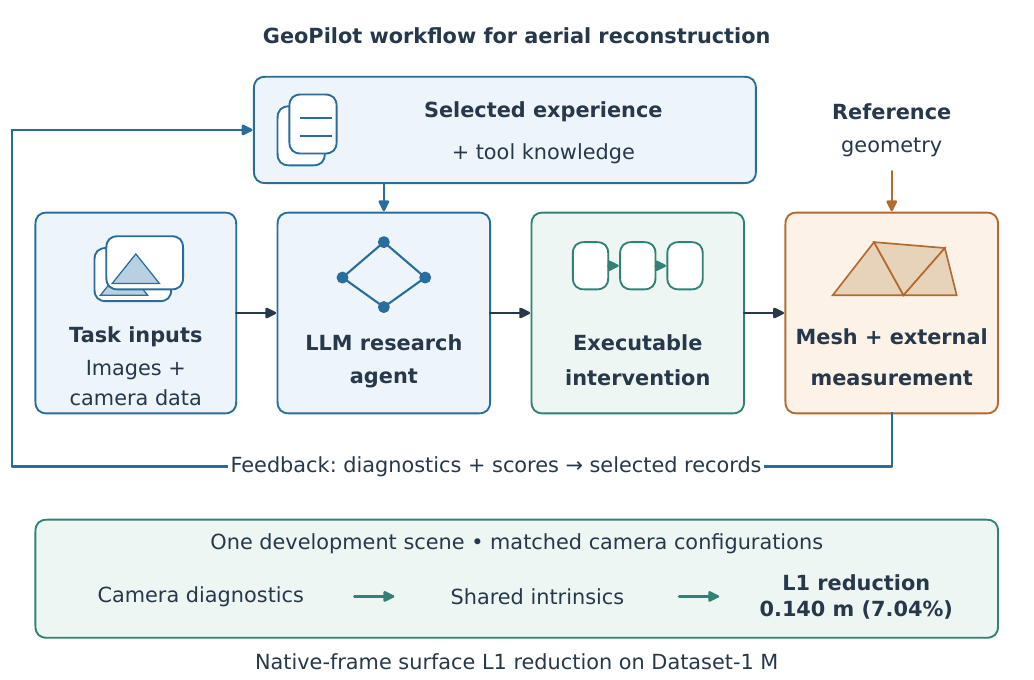}}
\caption{GeoPilot overview. Selected experience and tool descriptions inform diagnosis and executable reconstruction; numerical tools produce meshes for external evaluation. The lower strip reports the reduction in native-frame surface L1 for shared versus per-image intrinsics on Dataset-1 M (0.140 m, 7.04\%). Reference-assisted alignment reverses that ordering (Section~V-C), limiting a local-shape interpretation. Figure 2 details the two execution paths.}
\end{figure*}

Matched camera comparisons provide the main empirical cases. They test a diagnosis-guided configuration change and examine how its interpretation depends on coordinate treatment. Restricted-program experiments exercise the checked execution path. Recommendation and optional-review studies, including Jev \cite{ref3}, examine how prior observations and semantic checks affect subsequent decisions and complete-system costs.

GeoPilot changes numerical procedures and the external context supplied to the agent while keeping model weights fixed. We relate this distinction to recent work on recursive self-improvement \cite{ref4,ref5}.

\section{Related Work}\label{related-work}

\subsection{Reconstruction tools and geometric evaluation}\label{reconstruction-tools-and-geometric-evaluation}

GeoPilot uses COLMAP for incremental structure-from-motion (SfM) \cite{ref1} and OpenMVS for dense reconstruction, meshing and surface processing \cite{ref2}. DUSt3R estimates geometry through learned pointmaps and alignment \cite{ref6}, while numerical stereo methods adapt sampling and view selection \cite{ref7}. GeoPilot works at the level of reconstruction decisions: diagnostics guide the configuration and execution of these numerical components.

UseGeo already supports depth, point-cloud and triangle-mesh evaluation \cite{ref8}. Its published analysis includes calibration controls and identifies focal-length estimation as difficult for nearly nadir imagery. Those SfM results use reference-based alignment, including iterative closest point (ICP) with scale, whereas our principal measurements retain native coordinates. The numerical scores therefore answer different questions. Our protocol specifies area sampling, separate forward and reverse reference support, and artifact identities; it does not introduce a new acquisition dataset. DTU's structured-light references \cite{ref9} and Tanks and Temples' laser references \cite{ref10} further illustrate why acquisition conditions, reference support and coordinate treatment must accompany a reconstruction score.

\subsection{Configuration search and photogrammetric controls}\label{configuration-search-and-photogrammetric-controls}

A parameter-only program edit belongs to a configuration-search domain. Random search explores parameter combinations \cite{ref11}, while photogrammetric factorial studies examine calibration, bundle adjustment and camera-positioning settings \cite{ref12}. Shared and fixed camera intrinsics are established SfM controls, documented in COLMAP \cite{ref13}. GeoPilot examines how reconstruction diagnostics motivate a particular intervention and how its geometric consequences can be measured under matched inputs.

\subsection{Agent workflows, feedback and experience}\label{agent-workflows-feedback-and-experience}

ReAct interleaves reasoning, actions and observations \cite{ref14}; Reflexion reuses verbal feedback and episodic memory without weight updates \cite{ref15}. ExpeL extracts natural-language insights from successful and failed trajectories \cite{ref16}, while Memp studies the construction, retrieval and updating of procedural memory \cite{ref17}. GeoPilot supplies curated experience entries that distinguish observations, interpretations and applicability conditions. These entries were fixed before the recommendation experiment.

AstraLOD3 generates and revises building-specific Python/Blender procedures using images, calibrated cameras and SfM evidence \cite{ref18}. HiRS-Agent organizes remote-sensing tasks through manager and specialist roles with intermediate verification and trained policies \cite{ref19}. GeoPilot focuses on aerial reconstruction procedures and external experience. Reasoning occurs in the outer host, while its restricted numerical executor runs without online language-model calls.

Commercial spatial agents also organize reconstruction workflows. The SpatialMind user manual describes natural-language task planning, coordination of specialist capabilities, local tool execution and artifact lineage. Its aerial-reconstruction example combines image quality checks, bundle adjustment, optional control-point marking, reconstruction and reporting \cite{ref20}. These documented capabilities place task orchestration and result inspection within an agent interface. GeoPilot focuses on the diagnostic basis of specific program or camera interventions and the geometric conclusions supported by matched development comparisons.

Dream-RSI updates exploration-policy code using discovery histories \cite{ref4}, and MetaRSI distinguishes changes to data, harnesses and models \cite{ref5}. GeoPilot modifies checked reconstruction programs or separately versioned configurations and adapters; numerical tools produce and evaluate the resulting geometry.

\section{GeoPilot Method}\label{geopilot-method}

\subsection{Task and system overview}\label{task-and-system-overview}

GeoPilot takes permitted scene inputs, tool knowledge and development observations and returns an executable procedure, an execution trace and, when execution succeeds, a mesh. Figure 2 separates the research agent from the numerical executor. The agent reads diagnostics and development scores, proposes a hypothesis and arranges its test. It can introduce diagnostic code, a versioned configuration or an adapter. For the shared-intrinsics case, a Codex session and a diagnostic agent examined saved camera models and specified the comparison; fixed scripts then performed bundle adjustment, reconstruction and scoring.

The numerical executor runs checked programs without online model calls. Its fixed interface registers 14 parameters for densification, meshing and refinement, including resolution, view counts, support, decimation and face-area limits. P0, P1 and P2 use this interface with a fixed preparation stage. Intrinsic sharing was unavailable there, so the camera intervention uses the outer configuration path, with separately recorded capabilities and guarantees.

The recorded hypothesis preceded the A/B test, but the historical controller was not specified prospectively (Supplementary Appendices A.1 and A.6).

\begin{figure*}[!t]
\centering
\pandocbounded{\includegraphics[width=\linewidth,height=0.62\textheight,keepaspectratio]{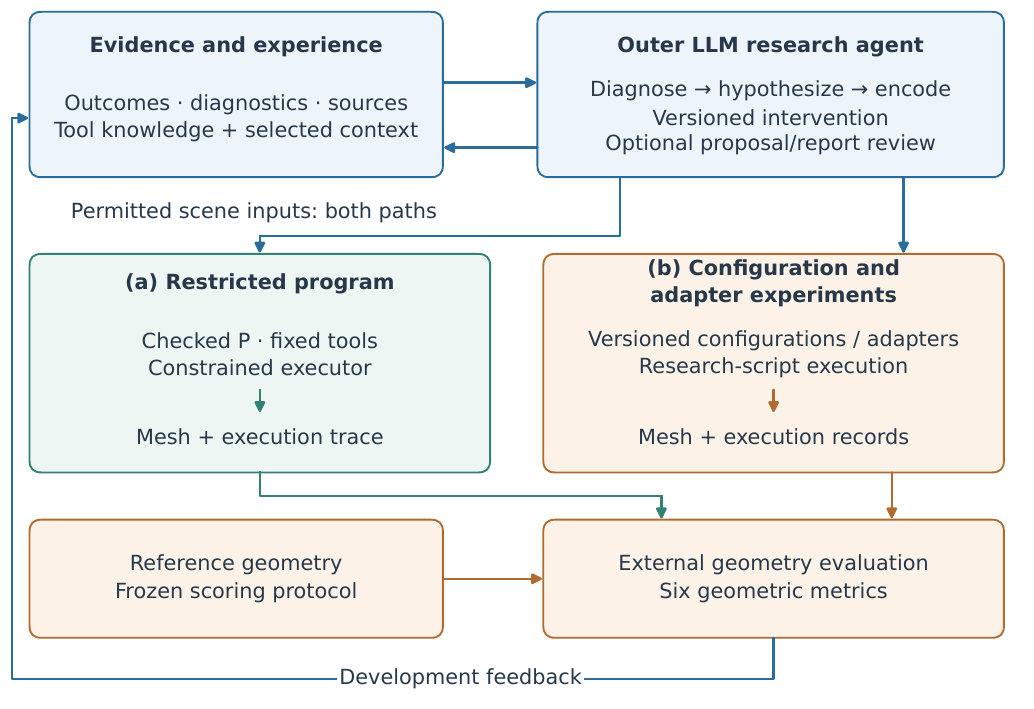}}
\caption{Two GeoPilot execution paths. The outer research agent can revise a checked program under fixed tools or introduce a versioned configuration or adapter experiment. Both produce artifacts for external evaluation. The shared-intrinsics case follows the configuration path. Experience selection and optional semantic review operate in the host.}
\end{figure*}

\subsection{Evidence and experience for diagnosis}\label{evidence-and-experience-for-diagnosis}

An experience entry records what changed, what was measured and which inputs produced the observation. Those input conditions determine whether the result is relevant to a later reconstruction. Entries distinguish expected effects from measured outcomes and preserve unresolved alternatives. Tool descriptions identify inherited settings and stage-specific parameters.

A history-building component supplies the proposer with earlier programs, their verified evaluation results and tool descriptions. P2 could read the P1 record while P0 remained the retained baseline. The component checks compatible parents and evaluation conditions, rejects duplicate candidate identities and preserves the intended history order. The outer workflow can additionally read diagnosis notes and unsuccessful research directions. In the camera case, poor reference agreement, camera diagnostics and homogeneous acquisition metadata motivated the shared-intrinsics hypothesis.

Experience entries enter the model context without changing its weights. Source versions and the context supplied to each proposal allow later decisions to be traced to earlier observations. The structured proposal interface in Supplementary Appendix A.1 records hypotheses, alternatives and predictions. It was not used in the shared-intrinsics experiment.

Camera experience is specific to the image/calibration relationship. Shared intrinsics on M concern calibration flexibility; fixed cameras on the publisher-undistorted raster track (U) concern using a supplied solution for those pixels. The Dataset-2 packet retains both observations with their distinct input conditions. The recommendation experiment records how this context is used on a subsequent task.

\subsection{Executable interventions}\label{executable-interventions}

For verified history \(H\), tool knowledge \(K\) and permitted edits \(\mathcal C\), the host-side proposer returns

\begin{equation}
(P',h,e,r)=F_{\theta}(P_0,H,K;\mathcal C),
\end{equation}

where \(h\) is the hypothesis, \(e\) contains expected observations and competing explanations, and \(r\) identifies supplied evidence. Model parameters \(\theta\) remain unchanged. An outer intervention instead returns versioned configuration or adapter code with an experiment plan. Program proposals operate within the registered numerical interface; outer configuration experiments can introduce additional preparation settings or adapters.

For permitted scene inputs \(x_s\), execution produces

\begin{equation}
(M_s,\tau_s)=\operatorname{Execute}(P,x_s;T,\omega_s),
\qquad M_s\in\mathcal M\cup\{\bot\},
\end{equation}

where \(T\) is the numerical interface, \(\tau_s\) the trace, \(\omega_s\) environment and numerical variation, and \(\bot\) failure to deliver a mesh. A checked program is a finite acyclic graph of tool calls, diagnostic checks and stop nodes. Validation checks registered parameters, stage requirements and all successors. Missing or nonfinite diagnostics follow an explicit unknown branch. Failed operations retain the preceding valid state and cannot be relabeled successful.

The restricted launcher binds the program, inputs, dependencies and candidate provenance. Its sandbox denies the numerical worker evaluation references, offline scores, personal configuration and network access. Actual calls and diagnostics are recorded. Before scoring, the delivered mesh is validated and its file hash is recorded to bind the evaluation to that artifact. The outer host performs separate numerical input checks without this sandbox. Supplementary Appendices A.1--A.3 and A.7 specify both execution paths and their reproduction requirements.

The restricted update can be summarized as follows; the camera-configuration path is outside this algorithm's edit domain.

\textbf{Algorithm 1. Restricted-program proposal, execution and feedback.}

\begin{enumerate}
\def\labelenumi{\arabic{enumi}.}
\tightlist
\item
  Read the stable parent's program, diagnostics and valid score; assemble compatible history, tool knowledge and the allowed edit domain.
\item
  Request a complete candidate program together with its hypothesis, expected observations, alternatives and evidence references.
\item
  Validate graph structure, stage requirements, numerical arguments and source bindings; execute only an accepted program, without rewriting its numerical content.
\item
  Record actual actions and observations on permitted inputs; validate the delivered mesh and record its file hash.
\item
  Evaluate the mesh under the parent's protocol; report every metric and difference, then apply the declared retention or promotion requirements.
\end{enumerate}

For example, P0 executes preparation, densification at resolution level 1, meshing with decimation 0.25, and stopping. P2 preserves that sequence and changes decimation to 0.5. The execution trace connects this single program edit to its numerical invocation and mesh. Conditional graphs support alternative paths, whereas the realized P0/P2 sequences are fixed paths. Full parameter bounds and structural checks are in Supplementary Appendix A.1. Supplementary Figure S1 distinguishes the executed program changes from the graph syntax.

\subsection{Geometric evaluation of interventions}\label{geometric-evaluation-of-interventions}

An external evaluator measures the mesh against protected reference data \(R_s\):

\begin{equation}
\mathbf q_s=E(M_s,R_s;\pi),
\end{equation}

where \(\pi\) fixes the protocol in Section~IV. All six measurements are retained; no composite reward or favorable subset is used to select an outcome. Feedback identifies the procedure that produced the measured mesh. Table~I summarizes the intervention surfaces and execution checks.

\begin{table*}[!t]
\centering
\caption{Intervention surfaces and execution checks.}
\label{tab:main-1}
\begin{tabular}{@{}
  >{\raggedright\arraybackslash}p{(\linewidth - 4\tabcolsep) * \real{0.3333}}
  >{\raggedright\arraybackslash}p{(\linewidth - 4\tabcolsep) * \real{0.3333}}
  >{\raggedright\arraybackslash}p{(\linewidth - 4\tabcolsep) * \real{0.3333}}@{}}
\toprule\noalign{}
\begin{minipage}[b]{\linewidth}\raggedright
Surface
\end{minipage} & \begin{minipage}[b]{\linewidth}\raggedright
Recorded content
\end{minipage} & \begin{minipage}[b]{\linewidth}\raggedright
Execution check
\end{minipage} \\
\midrule\noalign{}

Knowledge and experience & Sources, observations and applicability & Source identities and input compatibility \\
Diagnostics & Named measurements and artifact origins & Valid values and explicit unknown handling \\
Program structure & Calls, branches and stopping rules & Acyclic graph and stage requirements \\
Numerical arguments & Registered parameter values & Parameter types and bounds \\
Configurations and adapters & Camera settings, input bindings and source versions & Source identities and matched reconstruction settings \\
\bottomrule
\end{tabular}
\end{table*}

The six measurements answer different questions about the delivered surface. Mean L1 summarizes surface-to-reference distance; root mean square error (RMSE) gives more weight to large distances. Best-90\% summaries reveal the influence of the distance tail. Precision measures how much sampled mesh area lies within the threshold, while completeness measures how much reference support lies near a triangle surface. Reporting them together preserves these tradeoffs.

Before comparing outcomes, the evaluator verifies common protocol, reference support and input identities. Each evaluation record identifies the mesh and the executed procedure.

Preparation may fit a sparse reconstruction to permitted XYZ camera centers. Principal evaluation applies no fitted ICP, similarity correction, geometric repair or alignment to evaluation LiDAR. Products are assessed in their delivered native metric frame. Published camera inputs nevertheless have an upstream dependence: UseGeo describes joint image--LiDAR adjustment before release \cite{ref21}. Excluding reference files from local reconstruction does not make those supplied priors independent of LiDAR. The fixed-camera control consequently measures a complete available camera strategy, including its inherited information.

\subsection{Optional review and local information routing}\label{optional-review-and-local-information-routing}

The host can review a proposal or interpretation before accepting it. A reviewer checks evidence support and tool semantics using SUPPORTED, CONTRADICTED or INSUFFICIENT labels. Contradictions return to the proposer; unresolved execution prerequisites require evidence. An untested quality prediction can motivate an experiment whose configuration passes validation. Review neither produces a geometric score nor expands the restricted executor's numerical capabilities.

A separate routing configuration first filters experience by stage, then asks Jev or a full language model to select experience and a read-only diagnostic. The host reads the diagnostic source and sends the selected context to the common proposer. Unknown experience decisions conservatively retain the candidate set. Routing determines the context supplied to the proposer. Supplementary Appendices A.9--A.11 give the review and routing studies. The complete-system experiment in Section~IV-F tests whether savings in review-call time translate into shorter complete episodes.

\section{Experimental Protocol}\label{experimental-protocol}

\subsection{Data, research questions and study scope}\label{data-research-questions-and-study-scope}

The UseGeo-derived protocol contains three development scenes with 224, 327 and 277 images. Dataset-1 supplies the camera and program comparisons; Dataset-2 supplies the recommendation and complete-system studies. All three scenes participated in evaluator development. Table~II identifies the input, intervention and measured outcome for each comparison.

Two Dataset-1 raster tracks are distinguished throughout. M consists of 224 Metashape-directory JPEGs at \(7952\times5304\) pixels; U consists of 224 publisher-undistorted JPEGs at \(7953\times5279\) pixels. Corresponding filenames match, but every corresponding JPEG differs. The shared-intrinsics and P0/P1/P2 experiments use M. S-U/F-U uses U in both arms. Published calibration must therefore be bound to the actual raster rather than inferred from a filename.

Dataset-2 uses 327 U images with corresponding orientation records. An additional original JPEG lacks a matching orientation row and undistorted counterpart and is outside this fixed set. Dataset-1 has 105,922,149 full and 88,655,810 refined LiDAR points; Dataset-2 has 146,308,310 full and 125,035,236 refined points. The source acquisition occurred in April 2021 \cite{ref21}, without independently certified per-file timing or absolute reference accuracy for the local bundle.

Figure 3 shows the U input scenes and the supplied camera-center layouts used to organize their image sets. Photographs provide acquisition context; the camera locations are not ground footprints.

\begin{figure*}[!t]
\centering
\pandocbounded{\includegraphics[width=\linewidth,height=0.62\textheight,keepaspectratio]{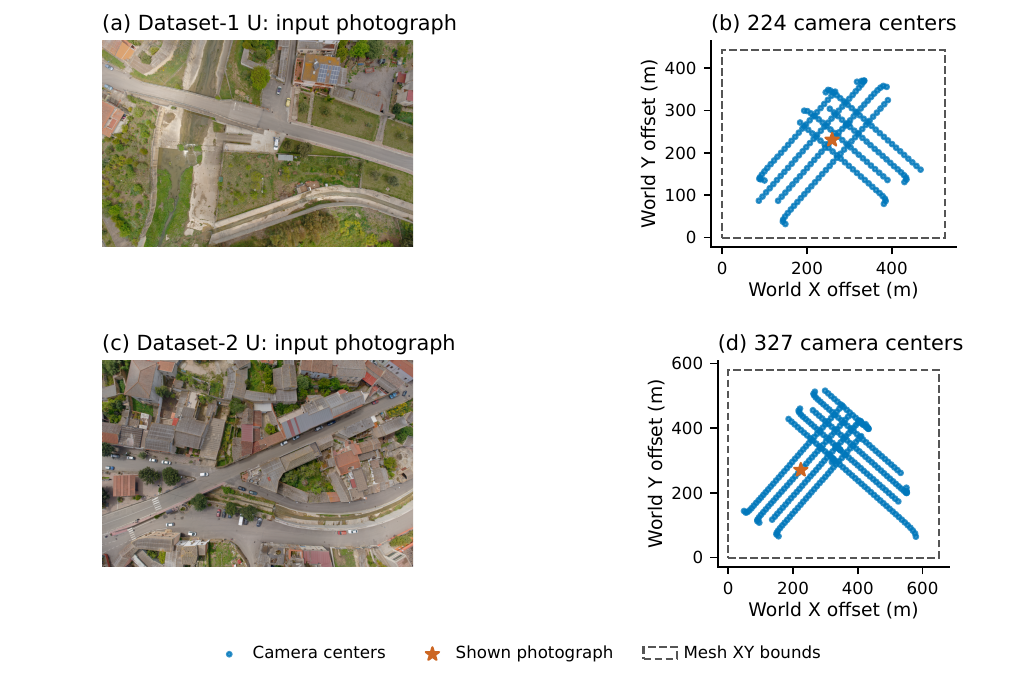}}
\caption{Input scenes and camera-center layouts for Dataset-1 U and Dataset-2 U. Each photograph is the middle image in filename order, selected independently of reconstruction scores. Maps show all supplied camera centers, identify the displayed photograph and outline the mesh XY bounds. These bounds are neither a visibility mask nor a ground footprint. The photographs are not overlaid on the maps. Source imagery: UseGeo \cite{ref21}.}
\end{figure*}

The labels distinguish inputs, configurations and procedures: M/U identify raster tracks; A/B identify the per-image/shared-intrinsics configurations on M; S-U/F-U identify estimated/supplied-camera strategies on U. P0/P1/P2 identify restricted programs, whereas R0/RF/RJ identify complete episodes with different review conditions.

\begin{table*}[!t]
\centering
\caption{Experimental design and measured outcome. M and U identify distinct raster inputs; all rows use development data.}
\label{tab:main-2}
\begin{tabular}{@{}
  >{\raggedright\arraybackslash}p{(\linewidth - 6\tabcolsep) * \real{0.2500}}
  >{\raggedright\arraybackslash}p{(\linewidth - 6\tabcolsep) * \real{0.2500}}
  >{\raggedright\arraybackslash}p{(\linewidth - 6\tabcolsep) * \real{0.2500}}
  >{\raggedright\arraybackslash}p{(\linewidth - 6\tabcolsep) * \real{0.2500}}@{}}
\toprule\noalign{}
\begin{minipage}[b]{\linewidth}\raggedright
Comparison
\end{minipage} & \begin{minipage}[b]{\linewidth}\raggedright
Input and unit
\end{minipage} & \begin{minipage}[b]{\linewidth}\raggedright
Change
\end{minipage} & \begin{minipage}[b]{\linewidth}\raggedright
Readout
\end{minipage} \\
\midrule\noalign{}

Camera A/B & Dataset-1 M; one mesh per arm & Per-image/shared intrinsics & Six metrics; proposal provenance; alignment sensitivity \\
S-U/F-U & Dataset-1 U; one mesh per arm & Estimated/fixed published cameras & Six metrics and image participation \\
P0/P1/P2 & Dataset-1 M; shared historical P0 & Restricted program edits & Complete vectors; retained parent \\
Experience & Dataset-2; three responses per condition & Direct, procedure, procedure plus experience & Nine recommendations and content checks \\
R0/RF/RJ & Dataset-2 U; one fresh mesh per arm & No added review/full-model/Jev review & Delivery, six metrics and complete elapsed time \\
\bottomrule
\end{tabular}
\end{table*}

\subsection{Geometry metrics and reference support}\label{geometry-metrics-and-reference-support}

For mesh \(M\), draw \(N=1{,}000{,}000\) samples uniformly with respect to triangle surface area, using NumPy PCG64 and frozen seed 20260916. With full discrete reference \(L\), define

\begin{equation}
d_i=\min_{\ell\in L}\lVert z_i-\ell\rVert_2,
\qquad z_i\sim\operatorname{UniformArea}(M),
\end{equation}

\begin{equation}
A_{L1}=\frac1N\sum_{i=1}^N d_i,
\qquad A_{RMSE}=\sqrt{\frac1N\sum_{i=1}^N d_i^2}.
\end{equation}

The same summaries over the 900,000 smallest distances, with index-stable ties, give best-90\% L1 and RMSE. At \(\delta=0.20\) m, precision is

\begin{equation}
\operatorname{Precision}_{\delta}=\frac1N\sum_{i=1}^N\mathbf1[d_i\leq\delta].
\end{equation}

Area sampling avoids weighting a region by mesh vertex density, although finite sampling still introduces approximation. Nearest-point errors also depend on reference spacing, missing observations and acquisition differences. The trimmed measures supplement the untrimmed errors and do not remove the distance tail from reporting.

Completeness uses the fixed refined reference \(L_r\) and distance to the triangle surface:

\begin{equation}
\begin{aligned}
c_j &= \min_{p\in M}\lVert\ell_j-p\rVert_2,\\
\operatorname{Completeness}_{\delta} &= \frac1{|L_r|}\sum_{\ell_j\in L_r}\mathbf1[c_j\leq\delta].
\end{aligned}
\end{equation}

Reverse queries use float64 point-to-triangle distances with bounded candidate materialization. Refined support follows the manifest's publisher-MVS-derived 5 m XY rule, fixed before candidate evaluation. It is a shared support convention, not an independently established visibility mask. Connected components, boundary and nonmanifold edges, degeneracies and surface area are separate mesh diagnostics. Supplementary Figure S2 illustrates why surface agreement and reference completeness require both query directions.

\subsection{Comparison units and numerical implementation}\label{comparison-units-and-numerical-implementation}

Each configuration delivers one reconstructed mesh. The million spatially related samples are evaluation queries on one mesh. Five-seed checks resample the same meshes; independent reconstruction replicates were unavailable. We report paired raw differences without reconstruction confidence intervals or significance tests. P1 and P2 share the historical P0 score, which is one baseline observation. Their preparation and densification were rerun, so matching upstream settings does not fix realized upstream geometry.

Principal scores use common six-metric definitions, reference support and native coordinates. Image participation is checked separately from geometric completeness. Supplementary Appendices A.3, A.5 and A.7 account for proposal attempts, reconstructions, scoring invocations and environment differences. Integration fixtures and unscored proposals are excluded from performance comparisons. Reproduction requirements and the curated release's distribution scope are given in Supplementary Appendix A.2.

\textbf{Numerical implementation.} Preparation uses CPU feature extraction and matching through pycolmap, with a SIMPLE\_RADIAL camera model and separate per-image camera entries. The inherited sparse configuration limits feature extraction images to 1600 pixels and features to 8192, uses sequential overlap 5, and adds spatial candidate pairs with at most 10 neighbors and a maximum distance of \(10^9\) m. Pair sets are united before incremental mapping. These preparation settings were fixed for the program comparisons. OpenMVS undistortion subsequently uses a separate 2400-pixel limit.

The sparse reconstruction is fitted to the permitted camera centers using a similarity transformation. At least three corresponding centers and nondegenerate centered geometry are required; the transform must have a finite positive scale and a proper rotation. The fit uses the permitted camera centers and has no access to the evaluator's LiDAR. The resulting transform is applied to mesh vertices on export, while OpenMVS continues to operate on its reconstruction scene.

The adapter fixes eight OpenMVS threads and archive type 2. Densification disables estimated-region cropping and tower mode; mesh construction disables region cropping and exports OBJ before conversion to the final PLY. Explicit program parameters are added to these fixed arguments. Unspecified parameters follow the executed tool's effective configuration. Stage-local settings do not necessarily control later operations: mesh-stage decimation and refinement-stage decimation belong to different numerical calls.

For OpenMVS 2.4.0 \cite{ref2}, \texttt{mesh.decimate=1} disables that decimation operation; cleanup, hole filling and smoothing have separate controls and are not disabled by this setting. The densification parameter \texttt{number-views-fuse} sets the minimum agreeing-image support during fusion, whereas \texttt{number-views} controls depth-map estimation. Changing fusion support can alter the point cloud supplied to subsequent mesh reconstruction and processing, so its effect cannot be interpreted simply as removing erroneous triangles.

\subsection{Camera and experience comparisons}\label{camera-and-experience-comparisons}

For M, A and B start from one sparse reconstruction and reset intrinsics to the same initialization. A retains per-image intrinsics; B shares one intrinsic model. Both fix the principal point and refine focal length, radial distortion, poses and sparse points with identical solver settings. Each receives fresh undistortion, densification at resolution level 1 and meshing with decimation 0.5. Georeferencing uses the same permitted XYZ centers.

For U, S-U and F-U share 224 published images, extracted features and matches. F-U fixes published calibration and poses during triangulation; S-U estimates cameras from images. Published XYZ centers support pair selection and subsequent georeferencing where applicable. Both receive fresh downstream reconstruction with identical settings (Figure 4). The undistorted-image limit is 2400 pixels, distinct from the 1600-pixel feature-extraction limit. The complete camera conversion, initialization and model-selection rules are in Supplementary Appendix A.4. Local camera-byte bindings are recorded; the precise publisher-release and coordinate mapping remains uncertified.

\begin{figure*}[!t]
\centering
\pandocbounded{\includegraphics[width=\linewidth,height=0.62\textheight,keepaspectratio]{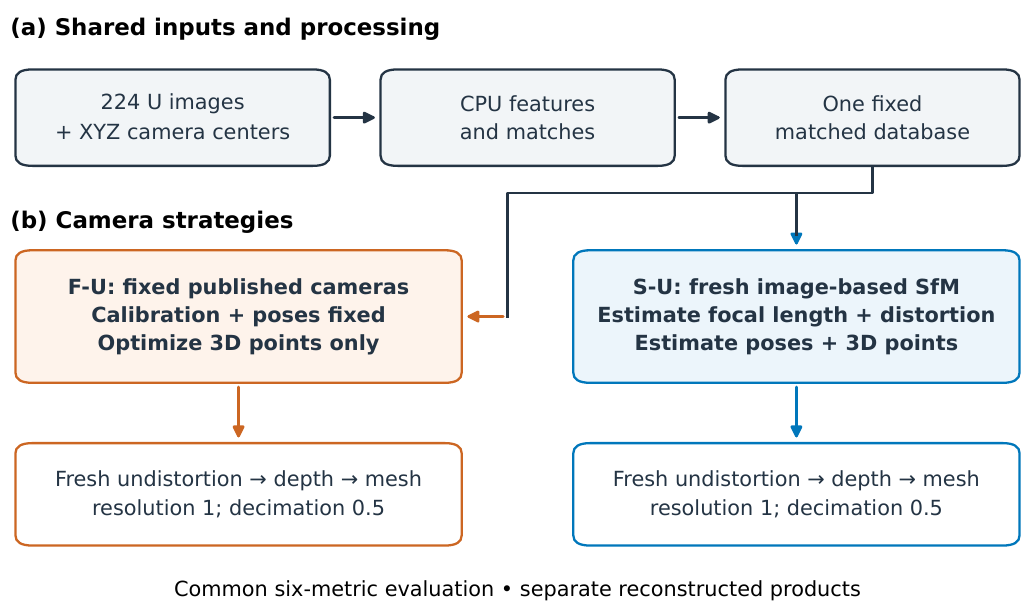}}
\caption{Camera-strategy control on identical U images and matches. F-U fixes published calibration and poses; S-U estimates cameras from images. Permitted XYZ centers support pair selection and georeferencing. Fresh downstream reconstruction uses common settings. Published cameras retain the upstream adjustment dependence described in Section~III-D.}
\end{figure*}

The Dataset-2 recommendation task was fixed before its F/S geometric outcomes were available. Direct analysis, the GeoPilot procedure without experience and the same procedure with experience received identical scene facts, tools and output requirements. Three fresh agent sessions per condition produced nine proposals, with randomized order within blocks. The available actions were F-U, S-U or a hold identifying missing information. No reconstruction score was associated with these recommendations; the subsequent end-to-end experiment was a separate study.

\subsection{Feedback-review validation}\label{feedback-review-validation}

We tested whether a reconstruction-specific rubric improves evidence checking and whether Jev reduces review cost relative to a full language model. Across five separately frozen development batches, 133 live API calls compared MiniMax-M3 (F) with Jev-1.13.0 plus MiniMax fallback (J). Each reviewer used generic (G) and task-specific (T) criteria and received the same evidence and three-way target: SUPPORTED, CONTRADICTED or INSUFFICIENT.

The generic rubric checks evidential sufficiency, internal consistency and executability while distinguishing observations from predictions. The task rubric adds effective tool and stage semantics, inherited settings, numerical predicates, raster/calibration/coordinate consistency, experience applicability, complete delivery, geometric measurements and the scope of causal or cross-scene claims. Label definitions are identical. Because the task instructions are longer, the comparison does not isolate domain content from prompt length.

J used one fallback after a call failure or an insufficient answer when a decision was required. The nine researcher-defined questions came from four related historical task families used during rubric development. Repeated evaluations and two-question diagnostics were not independent samples. The earlier 12-claim shadow pilot, which includes a deterministic-rule comparator, provides historical feasibility evidence (Supplementary Appendix A.9).

Each batch includes failures and fallbacks. Diagnostic reruns and repeated questions are reported separately, with reference labels checked against the underlying records by the researchers. Timing includes request preparation, calls, parsing, fallback and response saving, but excludes rubric construction and reconstruction. Supplementary Appendix A.9 gives the batch results and complete timing definitions.

\subsection{Complete-system review experiment}\label{complete-system-review-experiment}

One common host compared no added semantic review (R0), MiniMax review (RF) and Jev with the same fallback rule (RJ). Each used all 327 Dataset-2 U images, the same task history and experience, numerical tools, deterministic validation and stopping rule. The candidate domain contains F-U and S-U; the objective was a valid mesh and all six evaluation metrics, followed by a report of the results and their limitations. Parameter editing and image subsets are outside this experiment.

A saved seed fixed the order as R0, RJ, RF. Each run began in a fresh agent session and independently generated features, matches, sparse geometry, depth maps and a mesh. The configured model was gpt-6-astra with xhigh reasoning; the exact backend snapshot and sampling settings were unavailable. Tool histories confirmed that each session read only its assigned input. This restriction was imposed through instructions, without an operating-system sandbox. Remaining semantic claims are reviewed after common deterministic checks, and neither reviewer can alter geometric evaluation.

All numerical runs used seed 917, a 2400-pixel undistortion limit, dense resolution level 1, mesh decimation 0.5 and eight OpenMVS threads. A numerical reconstruction failure permits the other untried branch or a reasoned stop; scoring failure leaves the episode unscored. Complete elapsed time runs from the start of per-condition preparation until the final report was saved, including model decisions, review, coordination, waiting, reconstruction and scoring. Numerical and reviewer intervals are also retained separately. One descriptive block cannot separate provider, cache, thermal or numerical variation; its cost comparison concerns the observed complete episodes.

\section{Results}\label{results}

\subsection{Agent-led shared-intrinsics intervention}\label{agent-led-shared-intrinsics-intervention}

Under the per-image SIMPLE\_RADIAL preparation, camera-center residuals were meter-scale. Together with the homogeneous acquisition metadata, these diagnostics motivated a shared-intrinsics hypothesis, documented before bundle adjustment and dense reconstruction began (Supplementary Appendix A.6). The comparison retained separately estimated poses and scene points and used the matched design in Section~IV-D. Fresh dense reconstruction and meshing carried the camera change through to the delivered surface.

Figure 5 summarizes the post-intervention camera diagnostics and native-frame mesh L1. Focal-length estimates vary across the per-image models; B has one focal length because sharing imposes that constraint. Camera-center residuals improve for some images and worsen for others, while mean point reprojection error increases from 1.400 to 1.523 pixels. The camera diagnostics therefore do not establish uniform calibration improvement; the delivered mesh must be evaluated separately.

\begin{figure*}[!t]
\centering
\pandocbounded{\includegraphics[width=\linewidth,height=0.62\textheight,keepaspectratio]{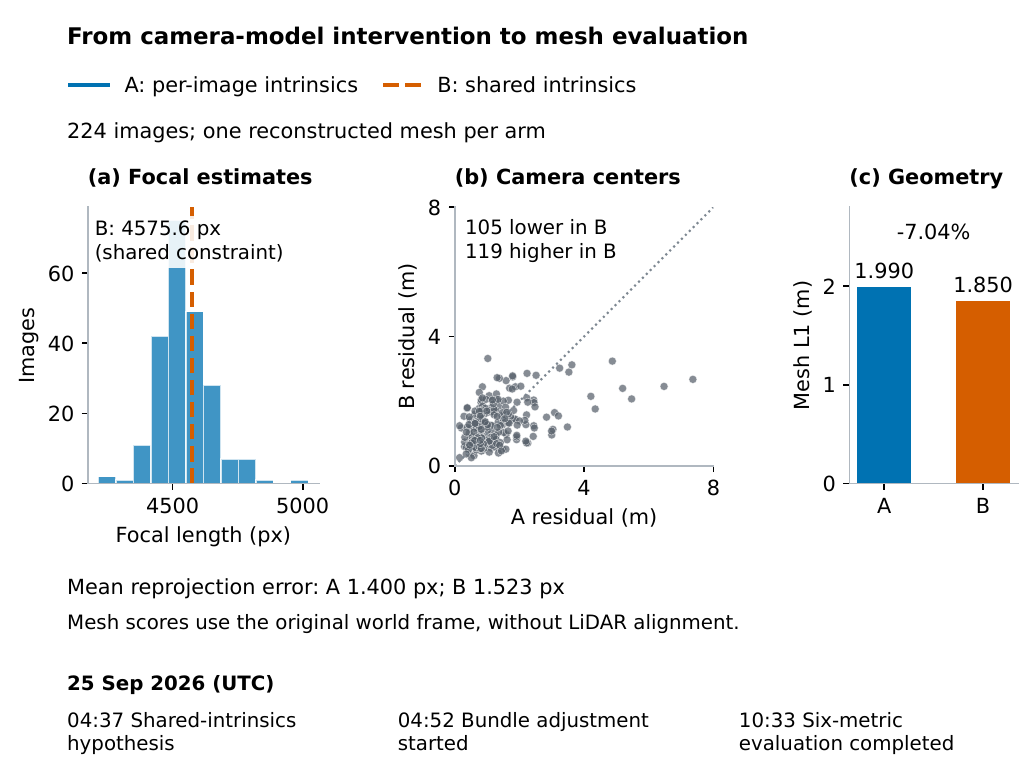}}
\caption{Camera diagnostics, shared-intrinsics intervention and native-frame evaluation on Dataset-1 M. Focal lengths and center residuals describe the models after the intervention; the single B focal value follows from parameter sharing. Center residuals are paired by the same 224 image names after each model is fitted to the permitted camera centers. The diagonal denotes equal residuals, not an accuracy target. Surface L1 uses the principal evaluation of one reconstructed mesh per setting; no reconstruction-repeat interval is available. The timeline places the hypothesis before matched reconstruction. Figure 9 gives the separate alignment-sensitivity diagnostic.}
\end{figure*}

All six measures favor shared intrinsics (Table~III). Surface L1 decreases by 0.139983 m (7.04\%) and RMSE by 0.131085 m (5.48\%); precision and completeness increase by 1.446200 and 0.704880 percentage points. The resulting L1 is 1.8497 m, with precision 7.8461\% and completeness 3.0695\% at 0.20 m. The gain improves agreement with the native-frame reference while substantial absolute residuals remain. Changes were calculated before rounding.

\begin{table*}[!t]
\centering
\caption{Intrinsic-sharing control on Dataset-1 M. A uses per-image intrinsics and B shared intrinsics; one complete reconstruction per configuration, scored in native coordinates. pp: percentage points.}
\label{tab:main-3}
\begin{tabular}{@{}
  >{\raggedright\arraybackslash}p{(\linewidth - 6\tabcolsep) * \real{0.4000}}
  >{\raggedleft\arraybackslash}p{(\linewidth - 6\tabcolsep) * \real{0.2000}}
  >{\raggedleft\arraybackslash}p{(\linewidth - 6\tabcolsep) * \real{0.2000}}
  >{\raggedleft\arraybackslash}p{(\linewidth - 6\tabcolsep) * \real{0.2000}}@{}}
\toprule\noalign{}
\begin{minipage}[b]{\linewidth}\raggedright
Metric
\end{minipage} & \begin{minipage}[b]{\linewidth}\raggedleft
Per-image A
\end{minipage} & \begin{minipage}[b]{\linewidth}\raggedleft
Shared B
\end{minipage} & \begin{minipage}[b]{\linewidth}\raggedleft
B \(-\) A
\end{minipage} \\
\midrule\noalign{}

Surface L1, m \(\downarrow\) & 1.9897 & 1.8497 & -0.1400 \\
Surface RMSE, m \(\downarrow\) & 2.3925 & 2.2614 & -0.1311 \\
Best-90\% L1, m \(\downarrow\) & 1.7101 & 1.5688 & -0.1413 \\
Best-90\% RMSE, m \(\downarrow\) & 2.0185 & 1.8767 & -0.1418 \\
Precision at 0.20 m, \% \(\uparrow\) & 6.3999 & 7.8461 & +1.4462 pp \\
Completeness at 0.20 m, \% \(\uparrow\) & 2.3646 & 3.0695 & +0.7049 pp \\
\bottomrule
\end{tabular}
\end{table*}

Five predefined sampling seeds, 20260925--20260929, preserve the favorable direction for all five surface-to-reference measures on the fixed meshes. Completeness is evaluated against every refined reference point. Section~V-C examines the coordinate sensitivity of the A/B mean-distance ranking.

\subsection{Fixed published cameras on undistorted images}\label{fixed-published-cameras-on-undistorted-images}

Both U strategies register all 224 images and produce 224 depth maps. F-U improves all six measures relative to S-U (Table~IV), with surface L1 lower by 0.024920 m (12.11\%) and precision and completeness higher by 5.875300 and 5.551237 percentage points. F-U additionally receives published orientations and retains the upstream image--LiDAR adjustment dependence; the contrast concerns complete camera strategies, including triangulation and coordinate interpretation.

Both strategies use the same features and matches, with 1766 candidate pairs from sequential and XYZ-neighbor rules. F-U optimizes triangulated points with published poses and intrinsics fixed. S-U estimates poses, focal lengths, radial distortion and points, holding principal points fixed; it receives no published orientations or F-U geometry. Both use permitted centers for pair selection and georeferencing where applicable.

Before OpenMVS import, the adapter reconciles image dimensions, intrinsics and two-dimensional observations while retaining poses and world geometry. Each branch then produces independent dense data and a mesh.

\begin{table*}[!t]
\centering
\caption{Camera-strategy control on Dataset-1 U. S-U estimates cameras and F-U fixes publisher cameras on identical undistorted images; one complete reconstruction per strategy. pp: percentage points.}
\label{tab:main-4}
\begin{tabular}{@{}
  >{\raggedright\arraybackslash}p{(\linewidth - 6\tabcolsep) * \real{0.4000}}
  >{\raggedleft\arraybackslash}p{(\linewidth - 6\tabcolsep) * \real{0.2000}}
  >{\raggedleft\arraybackslash}p{(\linewidth - 6\tabcolsep) * \real{0.2000}}
  >{\raggedleft\arraybackslash}p{(\linewidth - 6\tabcolsep) * \real{0.2000}}@{}}
\toprule\noalign{}
\begin{minipage}[b]{\linewidth}\raggedright
Metric
\end{minipage} & \begin{minipage}[b]{\linewidth}\raggedleft
Estimated S-U
\end{minipage} & \begin{minipage}[b]{\linewidth}\raggedleft
Fixed F-U
\end{minipage} & \begin{minipage}[b]{\linewidth}\raggedleft
F-U \(-\) S-U
\end{minipage} \\
\midrule\noalign{}

Surface L1, m \(\downarrow\) & 0.2059 & 0.1809 & -0.0249 \\
Surface RMSE, m \(\downarrow\) & 0.3359 & 0.3241 & -0.0118 \\
Best-90\% L1, m \(\downarrow\) & 0.1352 & 0.1087 & -0.0264 \\
Best-90\% RMSE, m \(\downarrow\) & 0.1743 & 0.1465 & -0.0279 \\
Precision at 0.20 m, \% \(\uparrow\) & 69.7435 & 75.6188 & +5.8753 pp \\
Completeness at 0.20 m, \% \(\uparrow\) & 66.8045 & 72.3557 & +5.5512 pp \\
\bottomrule
\end{tabular}
\end{table*}

Supplementary Figures S3 and S4 show how all six measures change within the M and U comparisons. Their absolute scores cannot rank the two interventions because the raster inputs and preparation differ. Figure 6 shows the S-U and F-U surfaces in common local windows; these renderings allow inspection of the geometry but do not measure reference error.

\begin{figure*}[!t]
\centering
\pandocbounded{\includegraphics[width=\linewidth,height=0.62\textheight,keepaspectratio]{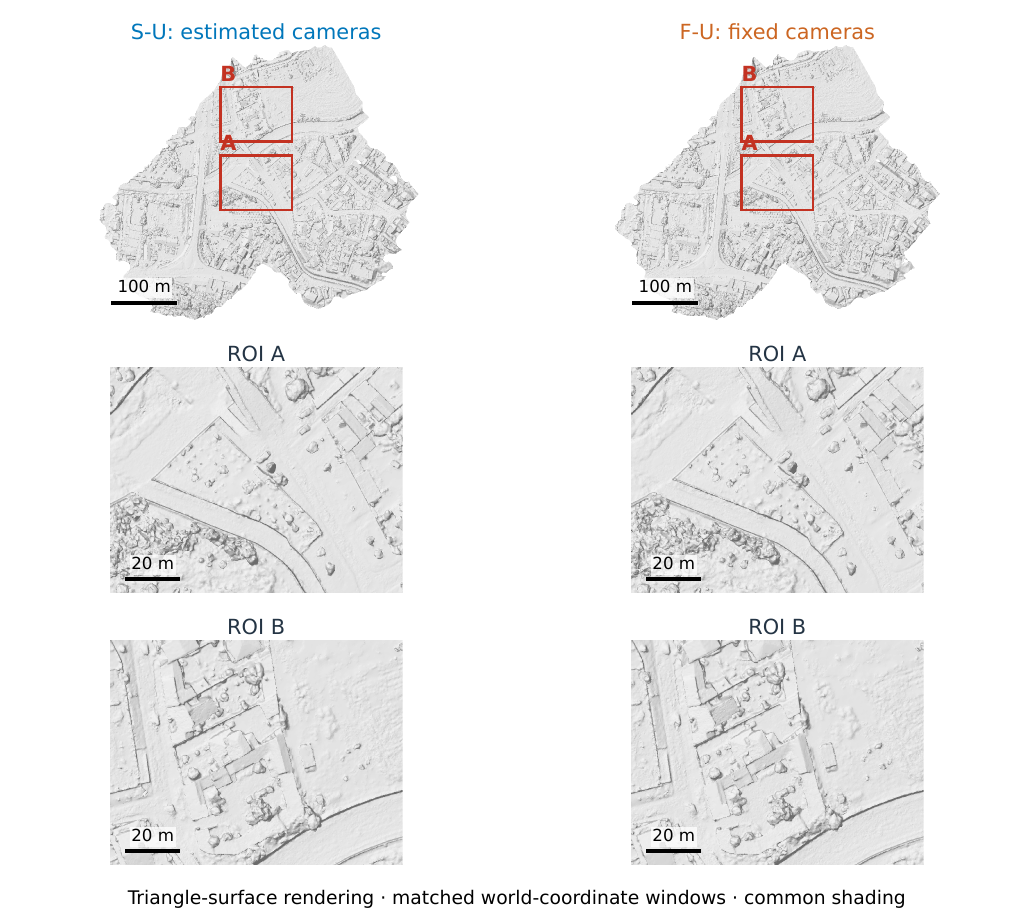}}
\caption{S-U and F-U mesh surfaces in a common top view and two matching local regions of interest (ROIs). All original triangles are rasterized with a depth buffer, identical view, scale and neutral lighting. The windows retain the same world-coordinate bounds in both products. Background pixels receive no visible triangle at the display resolution; the rendering does not measure completeness or local reference error. No surface repair, smoothing or geometric alignment is applied.}
\end{figure*}

\subsection{Spatial evidence for the shared-intrinsics case}\label{spatial-evidence-for-the-shared-intrinsics-case}

The spatial maps and alignment diagnostic address two ambiguities in the scene-wide scores: where the differences occur and how coordinate treatment affects the ranking. Figure 7 maps stored nearest-LiDAR distances for one million area-uniform samples per fixed mesh, with coordinates reproduced using the original sampling implementation and seed 20260925. This exploratory draw differs from Table~III (seed 20260916) and measures nearest-reference distance, not reverse completeness. The local window follows the joint mesh bounds and was not preregistered before the historical experiment.

Within the window, mean distance decreases from 3.009 m to 2.474 m (182,326 and 183,594 samples from the two meshes). Outside it, the mean decreases from 1.760 m to 1.707 m, while the 95th percentile increases from 3.924 m to 4.041 m. Thus the benefit is spatially uneven rather than a uniform correction.

More than 99.8\% of each mesh's sampled points lie in 1 m XY cells containing at least one full-reference LiDAR point. Empty horizontal reference cells alone therefore do not explain the widespread large distances in this display. That occupancy proxy does not establish vertically corresponding or visible reference surfaces, and unsigned nearest-point distances cannot separate global displacement, local shape error and reference mismatch. The remaining meter-scale distances require further diagnosis before a useful mapping-quality claim.

\begin{figure*}[!t]
\centering
\pandocbounded{\includegraphics[width=\linewidth,height=0.69\textheight,keepaspectratio]{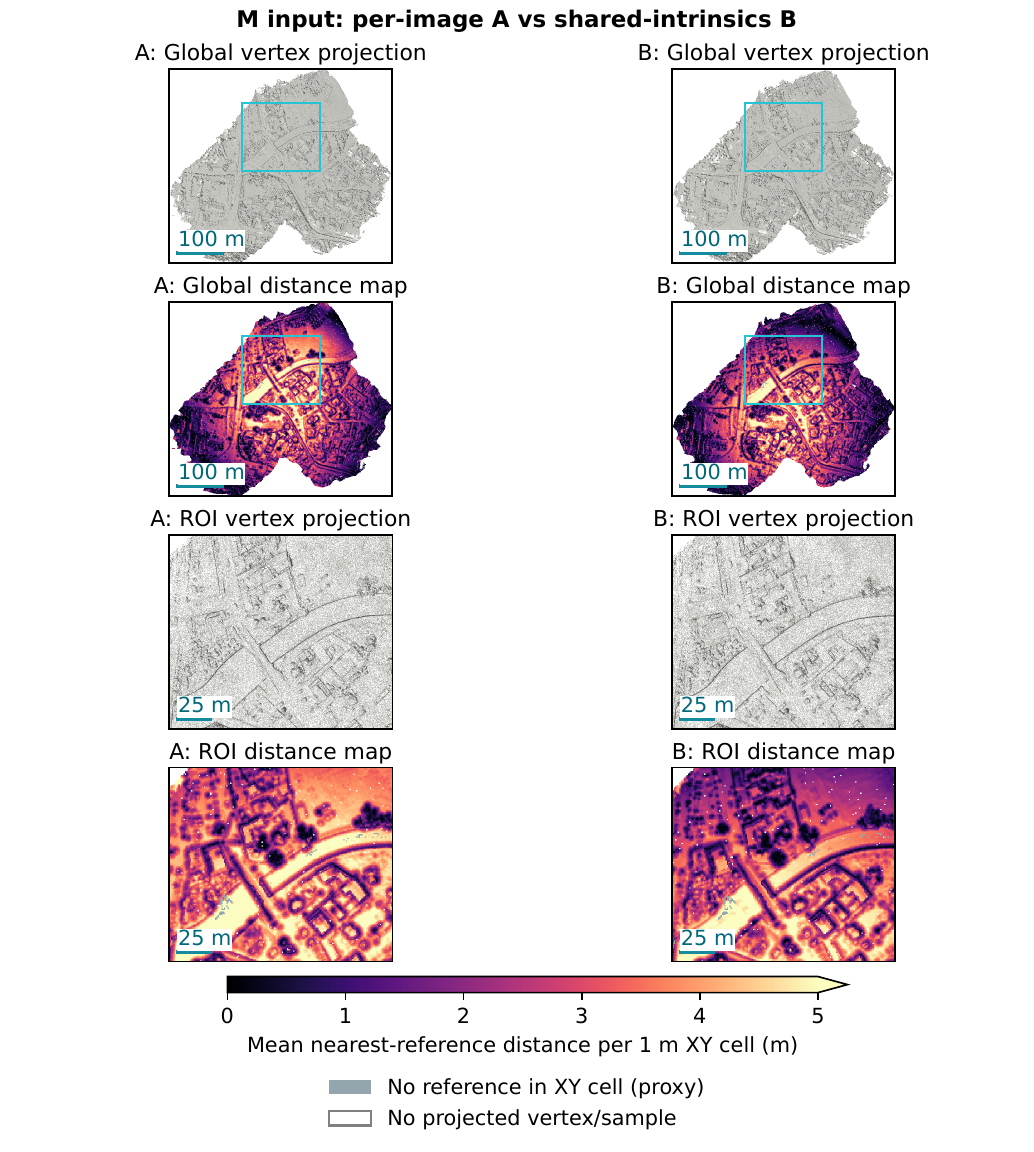}}
\caption{Shared-intrinsics A/B products in common top views and one local window. Rows show global vertex projections, global distance maps, ROI vertex projections and ROI distance maps. Gray panels project all vertices with common lighting; cyan boxes mark the same window. Distance panels give mean stored 3D nearest-reference distance per 1 m XY cell on a common 0--5 m scale, with larger values saturated. Gray-blue cells indicate mesh samples without full-LiDAR support in the same XY cell, a measure of horizontal occupancy rather than visibility. White marks missing projected vertices or samples, which alone cannot identify surface holes. Samples are not pointwise paired. The native-frame distance samples use seed 20260925; Table~III uses seed 20260916.}
\end{figure*}

Figure 8 complements the spatial maps with cumulative distance distributions for the full scene and the same window and its complement. Each curve describes area samples from a single fixed mesh. The distributions retain the full distance range and separate the two spatial subsets; they do not estimate uncertainty across reconstructions.

\begin{figure*}[!t]
\centering
\pandocbounded{\includegraphics[width=0.88\linewidth,height=0.62\textheight,keepaspectratio]{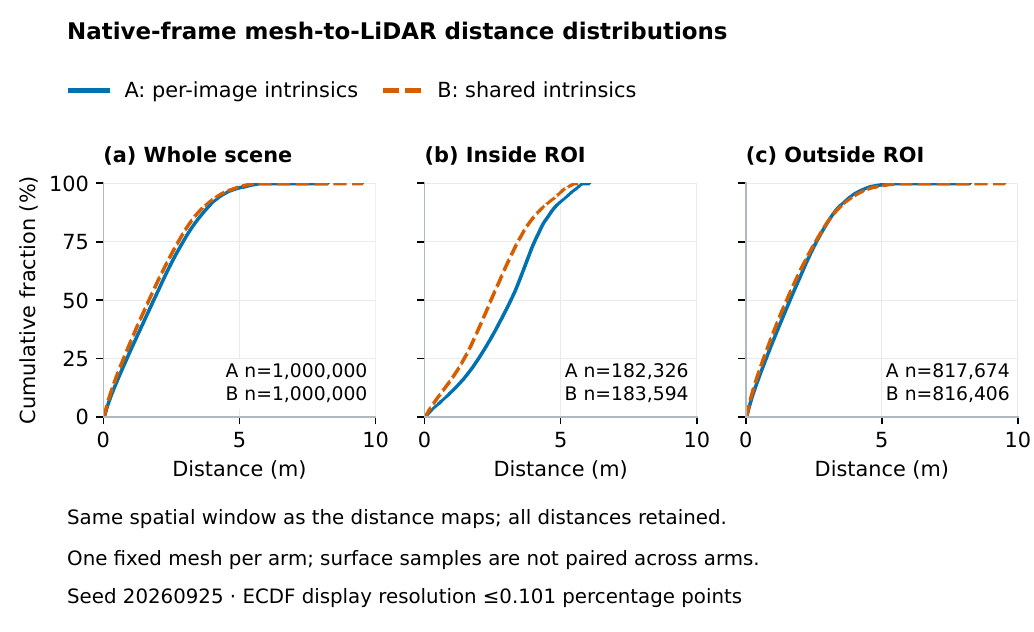}}
\caption{Empirical cumulative distributions of native-frame nearest-reference distance for A and B over the whole mesh, inside the window of Figure 7 and outside it. Each full-scene curve uses one million area-uniform samples with seed 20260925; subset counts are shown in the panels. All distances enter the distributions; curves are thinned by rank for display at a resolution of at most 0.101 percentage points. A and B are sampled separately from one fixed mesh per configuration, without pointwise pairing; the curves describe spatial variation, not variation across repeated reconstructions.}
\end{figure*}

The alignment diagnostic uses the same A/B area samples and the complete 105,922,149-point LiDAR reference. The sampling design was fixed before new distances were queried: 20 m world-XY checkerboard cells supply 6,000 fitting samples from even cells and 12,000 validation samples from odd cells, drawn separately for each mesh with seed 20260928. Identity-transform distances on the validation samples exactly match the stored values. Adjacent cells remain spatially related, so this is within-scene validation. The diagnostic performs no reconstruction or reverse-completeness evaluation.

Each translation, proper rigid and positive-scale similarity transform starts at identity and undergoes 20 nearest-neighbor iterations, fitting the closest 80\% of the fitting points at each iteration. Coordinates are centered before estimation; validation points never update the transform. Identity, translation, rigid and similarity results are all reported on the validation samples. The principal evaluation uses the unmodified meshes in their native frame.

Figure 9 compares validation mean distances for every transform. Translation, rigid and similarity fits all reverse the native-frame A/B ordering. After similarity fitting, mean distances remain approximately 0.96 m for A and 1.03 m for B. Thus the original benefit concerns delivery in the native coordinate frame and does not isolate improved local shape. These diagnostic fits are neither corrections applied to the reported meshes nor independent estimates of true coordinate offsets. Supplementary Appendix A.12 retains the exact values and transform details.

\begin{figure*}[!t]
\centering
\pandocbounded{\includegraphics[width=0.90\linewidth,height=0.62\textheight,keepaspectratio]{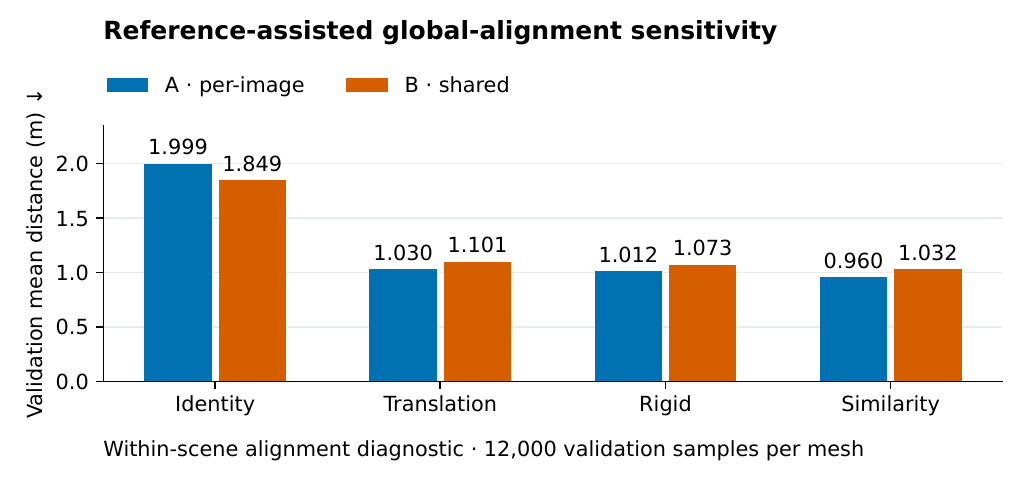}}
\caption{Fitted reference-assisted transforms reverse the native-frame A/B ordering. Bars show validation mean nearest-reference distance in meters for identity, translation, rigid and positive-scale similarity transforms. A uses per-image intrinsics; B shares intrinsics. Each mesh has 6,000 fitting and 12,000 validation samples drawn from alternate 20 m checkerboard cells; adjacent cells remain spatially related. These means describe within-scene alignment sensitivity; the principal scores use unmodified meshes, and reconstruction-repeat uncertainty is unavailable.}
\end{figure*}

\subsection{Restricted-program update}\label{restricted-program-update}

Neither restricted-program revision displaced P0. P1 changed conditional meshing and refinement, but all six measures favored its parent. P2 used the P1 history to change decimation from 0.25 to 0.5, increasing the mesh from 3,167,341 to 6,320,258 triangles. Its distance errors increased slightly, while precision and completeness changed by +0.003300 and +0.000497 percentage points. P0 was therefore retained. Fresh upstream reconstruction prevents attributing these differences to decimation alone (Supplementary Appendix B.1).

\subsection{Experience reuse in Dataset-2 recommendations}\label{experience-reuse-in-dataset-2-recommendations}

All nine Dataset-2 recommendations---three per context condition---selected F-U and passed configuration and content checks. Experience-conditioned responses reused earlier U evidence and distinguished it from M intrinsic sharing; direct responses made the same choice from current input compatibility. Table~V summarizes these outcomes; Supplementary Appendix B.2 gives all recommendation outcomes.

\begin{table*}[!t]
\centering
\caption{Dataset-2 U recommendation outcomes. Three fresh responses per context on the same task, with the same configured model and tools. Validity requires both configuration and content checks. These are recommendations without geometric scores.}
\label{tab:main-5}
\begin{tabular}{@{}lrr@{}}
\toprule\noalign{}
Context & Selected F-U & Valid responses \\
\midrule\noalign{}

Direct analysis & 3/3 & 3/3 \\
GeoPilot procedure without experience & 3/3 & 3/3 \\
GeoPilot procedure with M/U experience & 3/3 & 3/3 \\
\bottomrule
\end{tabular}
\end{table*}

A four-arm camera-decision study on four UAVScenes-derived development windows \cite{ref22} likewise converged on bundle adjustment in all 15 valid deliveries. The rule, generic model and experience-conditioned arms delivered 4/4, while the structured agent without experience delivered 3/4 because of a stop-schema error. These camera/track tests include no dense mesh score, and passing pixel checks did not certify metric geometry (Supplementary Appendix A.13).

\subsection{Feedback-review component measurements}\label{feedback-review-component-measurements}

Jev took less review-call time in the observed batches. In the nine-question batch with a clarified claim target, Jev plus fallback with the task rubric took 31.94 s, compared with 56.89 s for MiniMax; both agreed with seven reference labels but made different errors. These observations do not establish a stable benefit from the task-specific rubric or preserved judgment quality. Supplementary Table A4 reports the earlier and diagnostic batches separately, without pooling accuracy over 133 calls.

One failure illustrates why model fallback is not equivalent to obtaining evidence. A camera-numerics query asked whether the implementation reliably recognized coincident camera centers as a degenerate case. Jev initially abstained; the fallback model then accepted the incorrect claim. Supplying a direct execution fact corrected this query across the four reviewer--rubric conditions. Another query concerned the cause of a historical supervision failure: the available record did not distinguish competing explanations, yet the target-clarified task-specific Jev condition asserted a contradiction. The supervision-failure cause remained unresolved.

Supplementary Figure S5 reports all five batches and their individual query denominators; agreement is not pooled across repeated questions or diagnostic reruns. Supplementary Appendix A.9 gives failure accounting and the exact values.

\subsection{Complete-system development comparison}\label{complete-system-development-comparison}

All three conditions independently selected and completed F-U, with 327 images recorded at each processing stage, one numerical candidate, one complete score and no candidate failure or retry. Neither reviewer changed the proposed reconstruction action. Complete episode times were 89.81 min for R0, 93.35 min for RF and 102.23 min for RJ. Thus the observed RJ episode was 13.83\% longer than R0 and 9.52\% longer than RF. These are descriptive times for the common host implementation, including coordination and waiting time; they are not estimates of a stable hardware or deployment speed difference. Supplementary Table B4 reports every geometric measure, and Supplementary Figure S6 separates timing components; Supplementary Table A5 gives exact values.

Figure 10 illustrates the Dataset-2 delivery from R0, the condition without added semantic review; quantitative comparisons of review conditions are reported in Supplementary Table B4.

\begin{figure*}[!t]
\centering
\pandocbounded{\includegraphics[width=\linewidth,height=0.62\textheight,keepaspectratio]{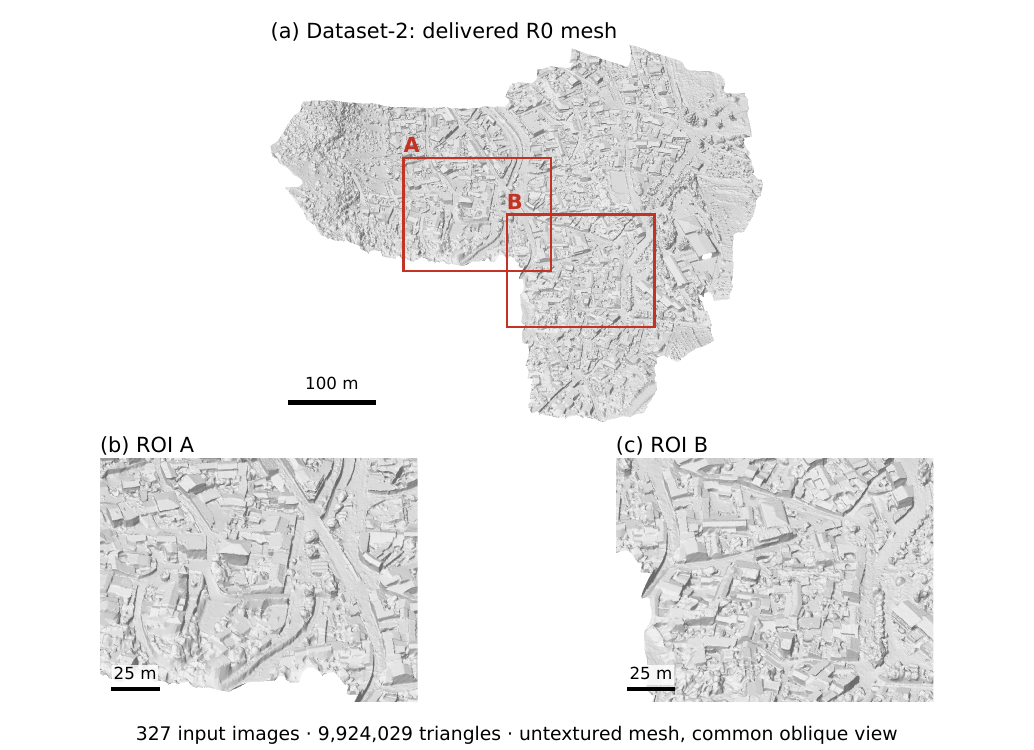}}
\caption{Delivered Dataset-2 U mesh from R0, reconstructed from all 327 input images using F-U. The mesh contains 9,924,029 triangles. The complete mesh is rendered in an orthographic oblique view, with two windows at fixed fractions of its projected extent and common neutral lighting. Window bounds and scale bars use projected view-plane meters. R0 is shown as the reference condition, independently of its appearance or score. This illustration establishes the delivered product, not a geometric advantage of any review condition.}
\end{figure*}

RF used two proposal-review calls; RJ used two proposal-review calls and one final-interpretation call. Neither used fallback, and all proposal reviews retained F-U. Only RJ required final-interpretation review under the frozen rule.

Jev returned INSUFFICIENT for the report's interpretation that larger-distance samples substantially affected the full-sample errors, and RJ left that interpretation unresolved. This model judgment added no geometric measurement. For the evaluated mesh, the largest-distance 10\% of samples contributed approximately 51.41\% of the total absolute distance and 94.78\% of the total squared distance. These proportions describe the tail's contribution to the recorded aggregate errors; they neither identify its cause nor constitute a significance test. The review changed the report's interpretation without changing the geometry.

Relative to R0, RJ's L1 is lower by 0.000750 m and precision higher by 0.129800 percentage points, while completeness is lower by 0.137221 percentage points. The other distance measures also decrease. Actions and settings are unchanged, so the mixed single-mesh differences cannot be attributed to review. The study establishes online review and full delivery under the recorded protocol, with no observed complete-episode acceleration.

\section{Discussion}\label{discussion}

\subsection{Camera handling as a reconstruction decision}\label{camera-handling-as-a-reconstruction-decision}

The camera cases return to the decision posed in the introduction: where should the next reconstruction change occur? On M, the diagnosis concerned calibration flexibility, so the matched test compared shared and per-image intrinsics. On U, the question was whether to reuse an available camera solution, so the comparison first required matching that solution to the pixels and coordinate conventions. Both interventions affected sparse geometry, undistortion and dense reconstruction before the mesh was evaluated. These observations complement the calibration sensitivity reported in UseGeo evaluation \cite{ref8}.

GeoPilot makes the proposed explanation testable by connecting it to an executable configuration and a matched geometric evaluation. That connection also constrains the conclusion. M improved in the native frame, but the fitted alignment diagnostic reversed its mean-distance ranking; U benefited from supplied cameras that retain upstream adjustment information. The results support different camera choices under different development inputs. Their absolute scores do not rank shared intrinsics against fixed cameras because M and U differ in pixels and camera information.

\subsection{Scope and limitations}\label{scope-and-limitations}

The camera gains concern matched development configurations, with one mesh per setting. Fixed-mesh resampling checks measurement variation rather than reconstruction uncertainty. The timestamped outer episode establishes a proposal preceding its outcome, but incomplete discovery-cost and backend metadata prevent estimating a discovery success rate.

Coordinate and reference assumptions bound the geometric interpretation. Reference-assisted fits reverse the M ordering, so its native-frame gain does not isolate local-shape improvement. Forward and reverse scores use different supports, the refined region is not a certified visibility mask, and supplied cameras retain upstream image--LiDAR adjustment dependence. The M residuals also remain too large to infer mapping suitability from the relative gain. Texture, radiometry and downstream application tolerances were not measured.

\subsection{Optional lightweight review}\label{optional-lightweight-review}

Faster review calls did not shorten the observed complete episodes. Reconstruction, scoring and coordination dominated their elapsed times, and neither reviewer changed the F-U action. The routing replay matched the rule arm's first diagnostic choices but still produced downstream interpretation errors. In one review case, a direct execution fact corrected a decision that model fallback had not resolved. This suggests that acquiring the missing evidence may be more useful than adding another semantic check.

\section{Conclusions}\label{conclusions}

GeoPilot makes a reconstruction diagnosis actionable: the agent proposes a program or camera-configuration change, numerical tools execute it, and an external evaluator measures the delivered mesh. The shared-intrinsics hypothesis preceded the matched comparison on Dataset-1 M, where the resulting reconstruction improved all six native-coordinate measures and reduced surface L1 by 7.04\%. A separate comparison on identical Dataset-1 U images favored fixed published cameras on all six measures, reducing L1 by 12.11\%. Three complete runs each delivered a mesh and full evaluation from all 327 Dataset-2 images.

The two camera results support different decisions. Shared intrinsics address calibration flexibility on M; fixed cameras reuse the supplied solution for U. The M gain concerns native-frame delivery, because the spatially split reference-assisted diagnostic reverses its mean-distance ranking. The U result retains the supplied cameras' upstream image--LiDAR adjustment dependence. With one reconstruction per configuration, these are conditional findings for the two development raster tracks.

\textbf{Data and code availability.} The access-controlled GeoPilot research archive (v0.1.0) contains selected source, curated knowledge and complete metric vectors with source identities for the camera and restricted-program comparisons (Tables~III--IV and B1--B2). These measurements support recomputation of those tables. Recommendation and review outcomes are described in Sections~V-E--V-G, with detailed tables and camera-decision outcomes in the supplementary material; their full execution records are locally archived and are outside the v0.1.0 archive. Geometric rescoring additionally requires the original meshes, reference data and frozen evaluation runtime; numerical reconstruction additionally requires raw rasters, compatible camera inputs and external binaries. UseGeo inputs must be obtained from the original providers under their terms; the \href{https://doi.org/10.1016/j.ophoto.2024.100070}{dataset publication} documents acquisition and release. Raw scene data, large reconstruction products, complete original evaluation records and the complete historical agent interaction and execution records are not redistributed. Exact agent replay is further limited by unavailable provider model snapshots. Original project code and curated original documentation use the MIT license; third-party files retain their notices. Supplementary Appendix A.2 distinguishes these reproduction levels.

\section*{Acknowledgment}
OpenAI Codex assisted with drafting and editing the abstract, Sections I--VII, figure captions, and supplementary text. Language models used within the research workflow are identified in Section III. The reported experimental measurements were retained from the recorded experiments. The authors are responsible for the manuscript and its scientific claims.
\onecolumn
\section*{Supplementary Material}
\setcounter{figure}{0}
\renewcommand{\thefigure}{S\arabic{figure}}
\renewcommand{\theHfigure}{supp.\arabic{figure}}
These appendices provide implementation details and extended results. Sections A.1--A.15 specify implementation and reproduction details; Sections B.1--B.3 provide extended program, recommendation and complete-system results. Unprefixed section and table numbers refer to the paper; tables prefixed A or B belong to these appendices.

\appendices
\setcounter{secnumdepth}{3}
\renewcommand{\thesubsection}{\thesection.\arabic{subsection}}
\renewcommand{\thesubsubsection}{\thesubsection.\arabic{subsubsection}}

\renewcommand{\thesubsectiondis}{\thesubsection.}
\renewcommand{\thesubsubsectiondis}{\thesubsubsection.}

\section{Program Contract and Reproduction}\label{program-contract-and-reproduction}

\subsection{Restricted program contract and executable instances}\label{restricted-program-contract-and-executable-instances}

Table A1 defines the legal parameter domain; the bounds are not optimized ranges. Preparation accepts no candidate parameters, and refinement requires a mesh. Because densification shares a depth-map workspace, the contract prevents repeating it along a path. A legal program may still produce poor geometry.

\Needspace{7\baselineskip}
\textbf{Table A1. Registered parameter ranges in the current program contract.} Integers and real-valued numbers are checked separately. Omitted parameters use the effective numerical-tool configuration.

\begin{longtable}[]{@{}llll@{}}
\toprule\noalign{}
Action & Parameter & Type & Allowed range \\
\midrule\noalign{}
\endfirsthead
\multicolumn{4}{l}{\small\itshape Table A1 (continued)}\\
\toprule\noalign{}
Action & Parameter & Type & Allowed range \\
\midrule\noalign{}
\endhead
\bottomrule\noalign{}
\endlastfoot
Densify & resolution-level & Integer & 0--2 \\
Densify & number-views & Integer & 2--20 \\
Densify & number-views-fuse & Integer & 2--5 \\
Densify & iters & Integer & 1--6 \\
Densify & geometric-iters & Integer & 1--4 \\
Densify & fusion-depth-diff-threshold & Number & 0.001--0.05 \\
Mesh & min-point-distance & Number & 0--5 \\
Mesh & free-space-support & Integer & 0--1 \\
Mesh & remove-spurious & Number & 0--40 \\
Mesh & close-holes & Integer & 0--60 \\
Mesh & smooth & Integer & 0--5 \\
Mesh & decimate & Number & 0.25--1 \\
Refine & resolution-level & Integer & 0--2 \\
Refine & max-face-area & Integer & 16--128 \\
\end{longtable}

A check node contains a metric name, comparison operator, threshold, and three explicit successors. A tool node contains an action, parameter object, and two explicit successors. A stop node has no operation. Duplicate identifiers, missing successors, unreachable nodes, cycles, unsupported fields, and unsafe numerical values are rejected. The 32-node and 16-tool-call limits apply to the program representation and an execution path. They are not a cumulative language-model-call budget and do not describe a scientific stopping rule for the research campaign.

In the initial conditional edit domain, satisfied and unsatisfied successors of a required check must have different normalized subgraphs. Renaming duplicate subgraphs does not qualify; structural difference does not prove that both conditions occur on the data. The later single-parameter domain permits a program without a condition.

The executor provides six diagnostics: registered-image fraction, sparse-point count, mean reprojection error, dense-point count, mesh-vertex count, and mesh-face count. Their definitions and artifact origins accompany the values. Online diagnostics use reconstruction artifacts; geometric quality is evaluated separately against LiDAR. The implementation permits graph reuse across scenes, but reuse is a representational capability; transfer performance has not been measured.

\textbf{Algorithm A1. Implemented restricted-program update.}

\begin{enumerate}
\def\labelenumi{\arabic{enumi}.}
\tightlist
\item
  Select the stable parent and read its actual program, diagnostics and valid development score.
\item
  Assemble the allowed actions, diagnostic definitions, tool knowledge and source-checked history. Keep each attempted child distinct from the stable parent and separate expectations from observations.
\item
  Request a complete program, hypothesis, expected effect, alternatives, risks and evidence references. For continuation, also request the proposed use of history and observable predictions under supporting and nonsupporting outcomes.
\item
  Validate program structure and the experiment's edit domain. Reject unsupported actions, illegal parameters, inconsistent source bindings or out-of-scope changes. Apply the accepted program without rewriting its numerical content.
\item
  Execute the program on permitted inputs, record the actions and observations, validate the delivered mesh, and record its file hash to identify the evaluated artifact.
\item
  Evaluate the valid mesh under the parent's frozen protocol and report all measurements and differences. Apply the declared promotion requirements; where these remain incomplete, retain the stable parent.
\end{enumerate}

The first recorded edit domain permits post-preparation changes to densification, meshing, refinement and structurally distinct conditions. For P2, the experiment design allowed exactly one registered mesh parameter to change while preserving all nodes, edges and other parameters. Historical provider settings and source bindings are reported in Appendix A.3.

In the execution mapping of Section~III-C, \(\bot\) denotes failure to deliver a mesh, and \(\omega_s\) includes environment and numerical variation. A program is a finite directed acyclic graph of tool, diagnostic-check and stop nodes. Tool nodes define parameters and success/failure successors; checks define a diagnostic, comparison, threshold and satisfied/unsatisfied/unknown successors. The executor maintains the current processing stage, artifact references, diagnostics and latest valid mesh. Structural validation propagates stage requirements across every successor, so an action must be legal on all paths that reach it.

These diagnostics are read directly from reconstruction artifacts. Missing, invalid or nonfinite values select the explicit unknown successor. These observations support executable decisions; their magnitude does not substitute for reference-based geometric quality. Failure retains the prior valid state and follows the declared failure successor. Stopping after a failed operation can return the preceding valid mesh, with the failure recorded. The numerical settings are given in Section~IV-C; Appendix A.7 specifies the environment and resources.

P0 executes \texttt{prepare\ -\textgreater{}\ densify(resolution-level=1)\ -\textgreater{}\ mesh(decimate=0.25)\ -\textgreater{}\ stop}; P2 changes only the decimation value to 0.5. Figure S1 places these edits alongside the later native proposals. All four displayed candidates realize fixed paths.

The later outer-agent interface associates each hypothesis with supporting observations, competing explanations, proposed actions and predictions. It classifies hypotheses as supported, contradicted, unresolved or untested and records whether prior experience applies to the current inputs. Each experiment also identifies its inputs, executed procedure, diagnostics, execution status and mesh measurements. This interface is separate from the historical camera episode.

\subsection{Reproducing the reported development comparison}\label{reproducing-the-reported-development-comparison}

The table-generation script reproduces the complete metric vectors in Tables~III--IV and Tables B1--B2 after checking score, protocol, reference and input identities. Appendix A.15 indexes selected local sources and P2 proposal checksums. The access-controlled GeoPilot research archive (v0.1.0) contains selected system code, evaluation utilities, instructions and curated knowledge. Its measurement file contains all six measures for P0, P1, P2 and the four camera-control products, with original score and manifest identities. The program-outcome record preserves interpretations of unpromoted attempts. Table recomputation uses these measurements; geometric rescoring and reconstruction additionally require separately acquired meshes, references, raw inputs and dependencies. Raw scene data, large reconstruction products, complete evaluation records and the complete agent interaction and execution records are not redistributed. Later recommendation, camera-decision and review results are reported in these appendices; their supporting execution records remain in the local archive and are not included in v0.1.0.

Table A2 distinguishes the public and local reproduction levels.

\Needspace{30\baselineskip}
\textbf{Table A2. Reproduction levels and additional requirements.} 

\begin{longtable}[]{@{}
  >{\raggedright\arraybackslash}p{(\linewidth - 4\tabcolsep) * \real{0.3333}}
  >{\raggedright\arraybackslash}p{(\linewidth - 4\tabcolsep) * \real{0.3333}}
  >{\raggedright\arraybackslash}p{(\linewidth - 4\tabcolsep) * \real{0.3333}}@{}}
\toprule\noalign{}
\begin{minipage}[b]{\linewidth}\raggedright
Level
\end{minipage} & \begin{minipage}[b]{\linewidth}\raggedright
Available in the research release
\end{minipage} & \begin{minipage}[b]{\linewidth}\raggedright
Additional requirements or limits
\end{minipage} \\
\midrule\noalign{}
\endfirsthead
\multicolumn{3}{l}{\small\itshape Table A2 (continued)}\\
\toprule\noalign{}
\begin{minipage}[b]{\linewidth}\raggedright
Level
\end{minipage} & \begin{minipage}[b]{\linewidth}\raggedright
Available in the research release
\end{minipage} & \begin{minipage}[b]{\linewidth}\raggedright
Additional requirements or limits
\end{minipage} \\
\midrule\noalign{}
\endhead
\bottomrule\noalign{}
\endlastfoot
Table recomputation & Complete vectors and original source identities for Tables~III--IV and B1--B2 & Recompute displayed differences from released measurements; complete original evaluation records remain in the local archive \\
Geometric rescoring & Evaluation utilities and protocol description & Original meshes, reference data and frozen runtime must be acquired separately \\
Numerical reconstruction & Selected source, numerical entry points and reproduction instructions & Raw rasters, compatible camera inputs, external binaries and experiment artifacts are required \\
Agent replay & Curated experience and selected workflow code & Complete historical agent interaction and execution records are not distributed; exact provider model snapshots are unavailable \\
\end{longtable}

\subsection{Proposal inputs and recorded cost}\label{proposal-inputs-and-recorded-cost}

P2 was proposed by MiniMax-M3 at temperature 0.2 with a maximum response allowance of 16,384 tokens. This per-call limit was not a campaign stopping rule. The provider did not expose the model-weight snapshot. The request contained a system instruction and one user message with the JSON-serialized development context: parent program, action registry, allowed diagnostics, execution limits, parent observations, ordered history, tool knowledge, edit plan and source bindings. Response-format adaptation reused the original proposal content without another model call.

The original request schema and checksums are listed in Table A9.

Tool knowledge included the software descriptions, effective settings and help excerpts. For component checksums, JSON was serialized with sorted object keys, preserved list order, UTF-8, unescaped Unicode and no separator whitespace; nonfinite numbers were rejected. Table A9 gives the resulting tool-knowledge and history checksums. Context construction and program validation sources are listed in Table A8.

The first conditional proposal and the P2 continuation have different edit contracts and knowledge inputs. They must therefore not be interpreted as an otherwise controlled history/no-history experiment. The saved request does not identify the provider's model weights or a generation seed. The v0.1.0 archive contains selected source and curated knowledge, not this complete historical request or every original execution dependency; Appendix A.2 states that scope.

The original provider response reports 13,073 tokens (8,024 input and 5,049 output) and approximately 44.01 seconds for the P2 proposal. These values cover that call, not the reconstruction or research campaign. Cost records list failed and interrupted attempts separately.

\subsection{Camera-control implementation and sources}\label{camera-control-implementation-and-sources}

A separate development control compares two complete camera-handling strategies on the same 224 publisher-undistorted images, denoted U. Both start from freshly extracted features and the same matched database, using the existing sequential and XYZ-neighbor rules, which yield 1766 distinct candidate image pairs. Feature extraction and matching use CPU execution, seed 917, and the resolution and feature limits in Section~IV-C. Each image has a separate SIMPLE\_RADIAL camera initialized with focal length 4636.912 pixels, principal point (3970.788, 2602.060) pixels and zero radial distortion. This initialization is tied to the \(7953 \times 5279\) raster; it is not transferred to the differently sized Metashape images.

In F-U, the published centers, orientations and calibration define fixed cameras. The adopted world-to-camera rotation is \(R_{cw}=\operatorname{diag}(1,-1,-1)(R_xR_yR_z)^T\), where the three rotations use the published omega, phi and kappa angles. The pixel-coordinate interpretation is \((c_x,c_y)=(x_0+0.5,-y_0+0.5)\); the values above already include that offset. Triangulation includes point optimization while camera poses and intrinsic parameters remain fixed. The numerical model uses the first image's camera center as a local origin; export adds that origin once. The local reconstruction never accesses the evaluation reference. The published camera information retains the upstream image/LiDAR adjustment dependence discussed in Section~III-D.

In S-U, a fresh incremental reconstruction starts from the same feature and matching input and initial intrinsics. It estimates camera poses, focal lengths, radial distortion and three-dimensional points, keeping principal points fixed. It receives no published orientation angles or F-U geometry. Camera centers enter shared spatial pair selection and the subsequent similarity fit for georeferencing, so S-U is not a reconstruction without geographic information. The predeclared model-selection order is registered-image count, then point count, then native model ID; it does not use reference scores.

Each selected sparse model receives independent undistortion, OpenMVS densification at resolution level 1 and meshing with decimation 0.5. The longest undistorted dimension is at most 2400 pixels. Image dimensions, intrinsics and two-dimensional observations are reconciled consistently before OpenMVS import, retaining poses and world geometry. No dense data or mesh is shared between F-U and S-U. Both meshes receive the complete six-metric evaluation. This comparison concerns the combined camera-handling strategies on U rather than an isolated orientation effect or restricted-program update.

The A/B and U camera comparisons each have separate reconstruction and evaluation files. Their numerical entry points and source locations are listed in Table A8.

Local manifests bind the camera-orientation bytes consumed by each run. The exact publisher release revision, horizontal CRS/EPSG identifier, vertical datum and authoritative orientation/pixel-origin mapping remain uncertified. The rotation and pixel formulas specify the implementation rather than certify those external fields. Score manifests bind meshes, protocol and sampling; source checks verify identities, complete reverse-query coverage and numerical transcription. These checks are distinct from independently recomputing every distance. The accompanying \path{paper/geopilot-camera-controls.csv} and \path{paper/geopilot-publisher-controls.csv} retain raw fractions and differences; tables convert fractions to percentages. Both pairs are development configuration comparisons.

\subsection{Native pilot design and provenance}\label{native-pilot-design-and-provenance}

A separate pilot used the Dataset-1 Metashape-raster input and permitted registered post-preparation edits while keeping preparation fixed. Two fresh Codex sessions each produced one proposal. Both received the same parent, action domain, diagnostic definitions, tool knowledge and task. Native-A received no outcome feedback; Native-B additionally received P0/P1/P2 programs, executed actions, diagnostics, costs and all six scores, together with a five-seed forward-scoring diagnostic comparing decimation settings on a fixed dense cloud. That diagnostic is neither repeated reconstruction nor a completeness measurement. Neither session received the camera-control outcomes. These native labels are distinct from the camera configurations A and B.

Native-A changed only mesh decimation from 0.25 to 1; Native-B changed only the fusion-support requirement from its default of 2 views to 3. Both retained a fixed path without diagnostic checks, so they did not test conditional selection. Each reconstruction started from fresh preparation. The evaluation design specified complete six-metric reporting, with untrimmed L1 as the primary descriptive endpoint. For valid deliveries with complete input-image coverage, a local Pareto improvement requires at least one strict improvement and no worsening of the other five measures; it does not establish a practically meaningful effect or justify promotion. One proposal per condition, different prompt lengths, unverified backend equivalence and fresh upstream reconstruction preclude estimating an average causal feedback effect. Table A3 summarizes delivery, artifact validation and scoring.

\Needspace{24\baselineskip}
\textbf{Table A3. Recorded native-pilot delivery and scoring status.} Native labels refer to Appendix A.5 and are distinct from the intrinsic-sharing configurations A/B. Historical P0 scores in Table B2 are not substituted for the contemporary controls.

\begin{longtable}[]{@{}
  >{\raggedright\arraybackslash}p{(\linewidth - 6\tabcolsep) * \real{0.2500}}
  >{\raggedright\arraybackslash}p{(\linewidth - 6\tabcolsep) * \real{0.2500}}
  >{\raggedright\arraybackslash}p{(\linewidth - 6\tabcolsep) * \real{0.2500}}
  >{\raggedright\arraybackslash}p{(\linewidth - 6\tabcolsep) * \real{0.2500}}@{}}
\toprule\noalign{}
\begin{minipage}[b]{\linewidth}\raggedright
Arm or attempt
\end{minipage} & \begin{minipage}[b]{\linewidth}\raggedright
Delivery status
\end{minipage} & \begin{minipage}[b]{\linewidth}\raggedright
Artifact validation
\end{minipage} & \begin{minipage}[b]{\linewidth}\raggedright
Complete six-metric result
\end{minipage} \\
\midrule\noalign{}
\endfirsthead
\multicolumn{4}{l}{\small\itshape Table A3 (continued)}\\
\toprule\noalign{}
\begin{minipage}[b]{\linewidth}\raggedright
Arm or attempt
\end{minipage} & \begin{minipage}[b]{\linewidth}\raggedright
Delivery status
\end{minipage} & \begin{minipage}[b]{\linewidth}\raggedright
Artifact validation
\end{minipage} & \begin{minipage}[b]{\linewidth}\raggedright
Complete six-metric result
\end{minipage} \\
\midrule\noalign{}
\endhead
\bottomrule\noalign{}
\endlastfoot
Native-A, original attempt & Completed and exported & Validated mesh & Unavailable; evaluation interrupted before completion \\
Native-B, original attempt & Failed supervision after numerical calls & Original execution remains failed & Absent \\
Native-B, artifact diagnostic & Recovered product only; no new reconstruction & Separate diagnostic submission valid & Not started \\
Contemporary P0, original attempt & Failed during dense fusion; no mesh & No valid mesh & Not applicable \\
Contemporary P0, additional serial attempt & Completed with revised supervisor & Validated mesh & Not started \\
\end{longtable}

Scheduling and supervisor versions differed across attempts. The pilot provides delivery and artifact observations without a complete comparative six-metric result. Recovery of the Native-B product does not change its execution-failure classification.

Both unchanged proposals were parsed with the same importer version, without requesting new responses. The provider did not expose backend model identity or token usage. Table A8 lists the original inputs, responses and execution files.

Native-A scoring stopped after 69,456,250 of 88,655,810 reverse queries, without a final score. Table A8 identifies the execution and scoring sources.

\subsection{Camera-intervention proposal provenance}\label{camera-intervention-proposal-provenance}

The camera experiment began with diagnosis of camera-center residuals at 04:07 UTC on 25 September 2026. The shared-intrinsics hypothesis was documented at 04:37, the matched-arm design at 04:43 and bundle adjustment began at 04:52. The first forward mesh measurements were obtained at 06:38, followed by the complete six-metric evaluation at 10:33. The hypothesis therefore preceded the matched bundle-adjustment and downstream reconstruction runs and their measurements. The later spatial residual analysis followed bundle adjustment.

Table A8 lists the diagnostic and planning files. The original camera-diagnosis session and its continuation are retained in the local archive and are outside the v0.1.0 archive.

The agent proposed and launched this camera-configuration experiment; fixed scripts performed bundle adjustment, reconstruction and scoring. Camera sharing was outside the restricted program's preparation interface. The available history establishes the observed sequence, with incomplete information on discovery cost and model calls.

\subsection{Numerical implementation, execution and environment}\label{numerical-implementation-execution-and-environment}

The preparation, georeferencing and OpenMVS execution settings are specified in Section~IV-C of this paper. The environment, artifact identities and resource safeguards are detailed below.

The reported measurements use \path{usegeo-mesh-geopilot-equal-v3-adaptive-batch}, a versioned implementation of the frozen protocol. P0, P1 and P2 share protocol, reference, input and sampling identities. Bounded-memory reverse-query scheduling was checked with synthetic examples, a complete small evaluation and a 100,000-query real-data probe; these tests cover the evaluated cases rather than establish all-scene bitwise equivalence. The reproduction index retains implementation identities and checks.

P0, P1 and P2 each considered all 224 input images for matching. All 224 images were registered and undistorted, and each was associated with a depth map. The adapter associates each depth-map header with a registered, undistorted input image. These counts measure image participation at each stage. They do not imply equal contributions to the mesh. Surface coverage is assessed separately by geometric completeness.

Each evaluation identifies the final mesh, program and scene inputs. P1 and P2 share the historical P0 reconstruction as their baseline, but rerun preparation and densification instead of inheriting a frozen dense cloud. Their dense-point counts can therefore differ even when settings match. Changing one program parameter limits the intended edit without holding upstream geometry constant.

Proposal calls, executable candidates, reconstructions and scoring invocations are counted separately. P1 required two provider calls for one executable candidate. The P2 response was converted to the required input format without an additional provider call or a change to its proposal content. Its first scoring attempt was interrupted; the later valid evaluation used the same mesh. Completed scoring durations exclude the interrupted attempt and do not represent total campaign cost.

All six measurements are retained under the same distance threshold, sampling protocol, reference support and coordinate treatment. Table B2 reports P2's small positive threshold differences alongside its increased distance errors.

The state and branch contracts are specified in Appendix A.1. Execution uses RSI-Harness commit \texttt{\seqsplit{33c4f8dfac4359987f2e814e187de67c332498de}}, with separately versioned local adapters and program contracts.

Reconstruction used Python 3.10.20 with NumPy 1.26.4, SciPy 1.11.1, and pycolmap 3.12.6. Evaluation used a separate Python 3.10.20 environment with NumPy 1.25.2, SciPy 1.11.1, Open3D 0.17.0, and laspy 2.7.0. The two environments therefore used different NumPy versions. Program execution also depends on the pinned RSI-Harness and local adapters described above, together with the bound OpenMVS binaries.

P0 and P1 evaluation manifests record macOS 26.5.2 on arm64; P2 records macOS 27.0 on arm64. Evaluator packages, protocol, references and sampling agree, but host platforms differ. Timing is descriptive, and platform-dependent numerical effects were not isolated.

For the historical program runs, the reconstruction supervisor allows four hours and 48 GiB of process-group resident memory. For the separate camera-configuration control in Section~V-A, each bundle-adjustment run has a 30-minute and 24 GiB process-group safeguard; each subsequent undistortion, densification, and meshing run has a separate 30-minute and 32 GiB safeguard. Both configurations complete within their stage-specific limits; these limits do not represent a combined campaign budget. Recorded supervisor durations use the implementation's monotonic clock and are distinguished from UTC timestamp differences. Evaluation has a separate four-hour watchdog, a 60 GiB resident-memory limit, and a recorded 0.5 s monitoring period. These are resource limits, not observed costs. The observed peaks remain in the per-run resource records with their measurement scope. A completed run within a limit does not establish feasibility on smaller machines, and no energy or monetary cost measurement is available.

\subsection{Recommendation-test design and records}\label{recommendation-test-design-and-records}

The Dataset-2 recommendation task was specified before its F/S reconstruction outcomes were available. It asked which camera strategy to attempt first: fixed publisher cameras (F-U), image-based estimation (S-U), or a hold identifying missing information. Direct analysis, the GeoPilot procedure without experience and the same procedure with experience received identical scene facts, tools, output requirements and evaluation objectives. The experience input contained the earlier Dataset-1 M/U results with their image-specific applicability conditions. Only this experience input differed between the latter two conditions.

Three fresh agent sessions per condition were run in three blocks, with order randomized within each block. Each session used gpt-6-astra with xhigh reasoning. The exact backend snapshot, generation seed, temperature and token usage were unavailable. Tool histories confirmed that each session read only its assigned input; this was an instructional restriction, without a platform-enforced sandbox. The nine original answers were evaluated without repair or outcome-dependent resampling. The readout included selected strategy, configuration validity, input applicability, comparison conditions and unsupported factual claims. F-U was not assigned a known-correct label.

This study evaluated recommendations; end-to-end reconstruction was tested separately. Its preflight covered inputs, runtime and sparse F/S stages. The later protocol in Section~IV-F adds downstream reconstruction and evaluation as a separate experiment. Dataset-2 remains development data in both studies.

All nine answers selected F-U and passed the configuration checks. Independent content review found no substantive violation. Experience citations revealed the context condition, so the reviewers were not completely blinded. Table A8 lists the task inputs, original responses and checks. These records postdate the v0.1.0 release.

\subsection{Feedback-review development sources and timing scope}\label{feedback-review-development-sources-and-timing-scope}

The preliminary review experiment contained four cases from a historical program family and eight from a synthetic decimation-semantics family. A thirteenth case requesting a geometric-benefit prediction was excluded before execution. Original and retrospectively augmented evidence were evaluated as separate conditions. Human extraction supplied the same structured predicates to both reviewers. Nine labels concerned verified parameter or documentation facts; the other three were provisional author judgments. Each model made one call per included case, without retries. Jev returned a typed choice and probabilities. MiniMax returned a label and reason with split reasoning requested; the saved response summary omitted the full HTTP response and separate reasoning trace. Jev timing included local request setup, while MiniMax used its provider-call timer. Both excluded rubric development, evidence preparation and reconstruction. The deterministic rule's agreement was recorded alongside model decisions. Table A8 lists the sources.

The subsequent factorial study is separate from this historical pilot. Both reviewers were asked for the same three-way decision, and all conditions shared the evidence for each query. Table A4 reports each batch separately, including incomplete responses and diagnostic reruns.

\Needspace{26\baselineskip}
\textbf{Table A4. Agreement with researcher-checked reference labels and summed per-query review time.} Each reviewer cell reports agreement/query count; seconds. F: MiniMax-M3; J: Jev with MiniMax fallback; G: generic rubric; T: task-specific rubric (Section~IV-E). Call counts include fallbacks and failed requests.

\begin{longtable}[]{@{}
  >{\raggedright\arraybackslash}p{(\linewidth - 10\tabcolsep) * \real{0.3258}}
  >{\raggedleft\arraybackslash}p{(\linewidth - 10\tabcolsep) * \real{0.1461}}
  >{\raggedleft\arraybackslash}p{(\linewidth - 10\tabcolsep) * \real{0.1461}}
  >{\raggedleft\arraybackslash}p{(\linewidth - 10\tabcolsep) * \real{0.1461}}
  >{\raggedleft\arraybackslash}p{(\linewidth - 10\tabcolsep) * \real{0.1461}}
  >{\raggedleft\arraybackslash}p{(\linewidth - 10\tabcolsep) * \real{0.0899}}@{}}
\toprule\noalign{}
\begin{minipage}[b]{\linewidth}\raggedright
Review condition
\end{minipage} & \begin{minipage}[b]{\linewidth}\raggedleft
F-G
\end{minipage} & \begin{minipage}[b]{\linewidth}\raggedleft
F-T
\end{minipage} & \begin{minipage}[b]{\linewidth}\raggedleft
J-G
\end{minipage} & \begin{minipage}[b]{\linewidth}\raggedleft
J-T
\end{minipage} & \begin{minipage}[b]{\linewidth}\raggedleft
Live calls
\end{minipage} \\
\midrule\noalign{}
\endfirsthead
\multicolumn{6}{l}{\small\itshape Table A4 (continued)}\\
\toprule\noalign{}
\begin{minipage}[b]{\linewidth}\raggedright
Review condition
\end{minipage} & \begin{minipage}[b]{\linewidth}\raggedleft
F-G
\end{minipage} & \begin{minipage}[b]{\linewidth}\raggedleft
F-T
\end{minipage} & \begin{minipage}[b]{\linewidth}\raggedleft
J-G
\end{minipage} & \begin{minipage}[b]{\linewidth}\raggedleft
J-T
\end{minipage} & \begin{minipage}[b]{\linewidth}\raggedleft
Live calls
\end{minipage} \\
\midrule\noalign{}
\endhead
\bottomrule\noalign{}
\endlastfoot
Nine questions, 4,096-token MiniMax response limit & 7/9; 67.10 & 7/9; 58.17 & 7/9; 22.79 & 7/9; 19.81 & 41 \\
Same nine questions, MiniMax client response limit removed & 7/9; 73.51 & 8/9; 54.54 & 7/9; 14.85 & 7/9; 20.13 & 38 \\
Two previously failed questions, additional execution evidence & 1/2; 25.23 & 2/2; 18.08 & 1/2; 1.69 & 1/2; 2.30 & 8 \\
Nine questions, clarified claim target & 8/9; 85.67 & 7/9; 56.89 & 7/9; 8.34 & 7/9; 31.94 & 38 \\
Same two questions, additional evidence and clarified target & 2/2; 25.64 & 2/2; 7.68 & 1/2; 1.04 & 1/2; 1.18 & 8 \\
\end{longtable}

The first batch retained three Jev HTTP failures and two incomplete MiniMax responses that reached the client cap; incomplete responses were not treated as semantic labels or removed from denominators. The rerun without the MiniMax client response limit used identical question, rubric and reference-label content. Subsequent diagnostic batches also retained the original outcomes.

The common clarification specified the top-level \texttt{state.claim} as the target and distinguished proposition truth from what a quoted report asserted; these two changes were not isolated. The two augmented questions were selected after errors were observed and were treated as diagnostic follow-ups.

No agreement counts are pooled across the five batches. Reference checking was internal and does not constitute external expert validation. The confusion tables report all three label classes, failed responses and unresolved decisions. These text experiments classified claims and performed no numerical reconstruction actions.

Review times include request preparation, calls, parsing, fallback and saving the responses, but exclude shared packet loading, rubric/evidence construction, annotation and numerical reconstruction or geometry scoring. Provider caching and concurrent host load were uncontrolled. Table A8 identifies the inputs and responses for the five batches, whose call counts sum to 133. Mock routing checks are excluded.

\subsection{Complete-system sources and resource accounting}\label{complete-system-sources-and-resource-accounting}

The three conditions were no added semantic review (R0), MiniMax review (RF), and Jev review with MiniMax fallback (RJ), as defined in Section~IV-F. Each run has separate reconstruction, score, input and execution records. The analysis verifies file identities and reconciles the timing components with total elapsed time. A separate integration execution and an interrupted dispatch that made no numerical or reviewer call were excluded. Table A8 identifies the protocol and analysis sources.

\Needspace{25\baselineskip}
\textbf{Table A5. Complete episode resources and component scope.} Minutes are elapsed host times. The review-and-coordination component includes time inside the review-call routine and revisions made during review. Routine time is reported separately in seconds as a subset of this component. Review-related preparation or reading outside the review-and-coordination interval contributes to residual host time, together with other agent and host waiting and record saving. The complete episode time includes all four components: reconstruction, geometric evaluation, review and coordination, and residual host time. First valid delivery is the time at which both a valid mesh and a full score were observed; it is not an additional component. Each condition produced one complete numerical candidate and one full score, and none used a fallback. Reviewer-call counts reflect the unresolved assertions encountered during each run.

\begin{longtable}[]{@{}lrrr@{}}
\toprule\noalign{}
Time or count & R0 & RF & RJ \\
\midrule\noalign{}
\endfirsthead
\multicolumn{4}{l}{\small\itshape Table A5 (continued)}\\
\toprule\noalign{}
Time or count & R0 & RF & RJ \\
\midrule\noalign{}
\endhead
\bottomrule\noalign{}
\endlastfoot
Complete episode, min & 89.81 & 93.35 & 102.23 \\
First valid delivery observed, min & 85.41 & 89.66 & 91.76 \\
Reconstruction, min & 47.70 & 52.55 & 53.18 \\
Geometric evaluation, min & 32.30 & 30.26 & 32.55 \\
Review and coordination, min & 0.00 & 0.81 & 8.76 \\
Time inside the review-call routine, s & 0.000 & 4.371 & 1.494 \\
Residual host time, min & 9.81 & 9.72 & 7.74 \\
Reviewer calls & 0 & 2 & 3 \\
Native turns & 3 & 3 & 4 \\
\end{longtable}

Provider metadata reports 161,614 tokens for R0, 160,814 for RF and 208,849 for RJ. Input/output counts are 157,687/3,927, 157,367/3,447 and 202,848/6,001; cached-input counts are 148,224, 109,184 and 184,064. These provider counts are reported separately from numerical compute and elapsed time. R0 token counts were obtained from the additional provider usage record. Coordinating researcher usage, implementation, rubric construction and annotation costs remain unmeasured.

Online review made five external API calls, separate from the 133 component-study calls. For RJ, the final-interpretation review returned INSUFFICIENT, and the revised response marked the interpretation as unresolved (Section~V-G). Both versions are retained in the archive. Input-file checks confirmed that the reviewed text matched the evaluated response and that each session read the assigned input.

\subsection{Active experience selection and diagnostic routing}\label{active-experience-selection-and-diagnostic-routing}

A separate exploratory replay used four researcher-reconstructed historical development states: image/calibration binding, dispersed estimated intrinsics, mesh/refinement settings and mixed candidate scores. Each state was run once under conservative rules, MiniMax-M3 routing and Jev-1.13.0 routing, in a frozen within-state order. All arms shared candidate knowledge, facts, diagnostic definitions and a MiniMax-M3 downstream proposer. The rule arm retained all stage-compatible experience; model arms selected context and one diagnostic in a single request. The selected experience was included in the downstream request, and diagnostics were read or recomputed from the identified source files. This was a read-only information-routing replay, not a new reconstruction or independent-scene test.

Across the four states, complete local elapsed times were 78.091 s for rules, 120.881 s for MiniMax routing and 36.188 s for Jev routing; routing itself accounted for 0.008, 47.569 and 2.153 s, respectively. The response summary records 20 HTTP calls: four downstream proposer calls in the rule arm, and four routing plus four downstream calls in each model-routing arm. Downstream user-packet bytes totaled 16,475, 10,473 and 11,257; including routing, serialized request-file bytes totaled 20,841, 40,125 and 41,167. Bytes are not tokens or wire traffic. Jev reported 6,900 input and 1,005 output tokens for routing; its common MiniMax downstream calls separately reported 3,981 prompt and 3,744 completion tokens. Provider tokenizers are not treated as one compute unit. The initial batch made 21 calls and included parser failures. The subsequent 20-call batch repeated the same twelve conditions with a parser that also accepted a response consisting of one complete JSON code block. The batches are reported separately because they repeat the same four development states, not twenty-four independent tasks.

In the subsequent batch, all twelve state-by-arm conditions returned schema-valid downstream proposals. Both model-routing arms selected the rule arm's source-backed first diagnostic in every state. MiniMax omitted the experience item on missing, interrupted or unresolved evaluations (SCORE\_STATUS) in one state, while Jev omitted the item on sharing camera intrinsics (SHARING) in another. These relevance annotations were defined by the researchers; score-status information remained available through a diagnostic. Content checks found downstream stage and diagnostic-capability errors, including attributing camera intrinsics to mesh/refinement settings. The common MiniMax proposer generated every downstream answer, and none of its follow-up interventions was executed. The replay measured context selection and local cost. It included no subsequent reconstruction. Sources are listed in Table A8.

\subsection{Reference-assisted global-alignment sensitivity}\label{reference-assisted-global-alignment-sensitivity}

Section~V-C of this paper gives the spatially split fitting/validation design and Figure 9 shows every tested transform. Table A6 reports the complete validation means; additional transform details follow the table.

\Needspace{14\baselineskip}
\textbf{Table A6. Diagnostic validation mean nearest-reference distances (m).} These 12,000-point spatial subsets differ from Table~III's sampling population and seed. The reference-assisted rows evaluate transforms fitted to LiDAR; the principal reconstruction scores in Table~III use unmodified meshes in native coordinates.

\begin{longtable}[]{@{}lrr@{}}
\toprule\noalign{}
Transform & A: per-image intrinsics & B: shared intrinsics \\
\midrule\noalign{}
\endfirsthead
\multicolumn{3}{l}{\small\itshape Table A6 (continued)}\\
\toprule\noalign{}
Transform & A: per-image intrinsics & B: shared intrinsics \\
\midrule\noalign{}
\endhead
\bottomrule\noalign{}
\endlastfoot
Identity & 1.999156 & 1.848829 \\
Translation & 1.030263 & 1.100730 \\
Rigid & 1.011898 & 1.073042 \\
Similarity & 0.959510 & 1.032006 \\
\end{longtable}

The translation estimates have Z components of -3.526 m (A) and -3.290 m (B), but are fitted nearest-neighbor displacements, not independently measured datum offsets. Mean errors remain approximately 0.96--1.03 m even after the tested similarity fits. Fixed iterations do not guarantee convergence or a global optimum; corresponding surfaces may be mismatched. The reversal of A/B ordering limits a shape-improvement interpretation of the original gain, but does not establish that A has intrinsically better shape. Per-cell summaries, final transforms and per-iteration fitting-distance summaries are retained in the diagnostic record. Neither reconstruction-repeat uncertainty nor reference systematic error is estimated by this analysis.

\subsection{Four-arm camera-decision development study}\label{four-arm-camera-decision-development-study}

A subsequent study used four eight-image windows from the HKisland01 and AMtown01 sequences of UAVScenes \cite{ref22}. Window membership was fixed from the first frame at or after 25\% and 75\% of each sequence's elapsed time, followed by seven images, before accessing their pixels. These are development tasks from two sequences. All arms received the same given-intrinsic camera inputs, image/track access and fixed-camera versus bundle-adjusted candidates. Whole tracks were separated into training and validation. Validation triangulated each admitted track from its first two views and measured prediction errors in the remaining views; it did not measure metric surface accuracy.

The deterministic arm R attempted the fixed-camera and BA alternatives and applied the frozen qualification and dominance rule. L was a generic tool-using model; G0 required structured decision-state updates; GE additionally received a five-entry historical experience snapshot and recorded its applicability. State tools were also available to L, so G0--L does not isolate the presence of feedback or state access. All model responses reported the identity MiniMax-M3.1-Flash-Preview. Each task--arm pair ran once, in a predetermined order. The practical development gate required sufficient training support and validation tracks, held-out median/p90 errors no greater than 2/5 pixels, positive-depth fraction at least 0.95 and median triangulation angle at least 0.5 degrees; a BA candidate also required a usable solver result. These checks do not certify depth accuracy.

\Needspace{18\baselineskip}
\textbf{Table A7. Camera-decision development outcomes and measured episode costs.} Costs are sums across four tasks, including the failed episode; common preparation is accounted separately. A fixed candidate materializes cached preparation, whereas each BA attempt performs a fresh training-only optimization. No dense reconstruction or final mesh evaluation is included.

\begin{longtable}[]{@{}
  >{\raggedright\arraybackslash}p{(\linewidth - 10\tabcolsep) * \real{0.1304}}
  >{\raggedleft\arraybackslash}p{(\linewidth - 10\tabcolsep) * \real{0.1739}}
  >{\raggedleft\arraybackslash}p{(\linewidth - 10\tabcolsep) * \real{0.1739}}
  >{\raggedleft\arraybackslash}p{(\linewidth - 10\tabcolsep) * \real{0.1739}}
  >{\raggedleft\arraybackslash}p{(\linewidth - 10\tabcolsep) * \real{0.1739}}
  >{\raggedleft\arraybackslash}p{(\linewidth - 10\tabcolsep) * \real{0.1739}}@{}}
\toprule\noalign{}
\begin{minipage}[b]{\linewidth}\raggedright
Arm
\end{minipage} & \begin{minipage}[b]{\linewidth}\raggedleft
Valid deliveries
\end{minipage} & \begin{minipage}[b]{\linewidth}\raggedleft
Fresh BA attempts
\end{minipage} & \begin{minipage}[b]{\linewidth}\raggedleft
API calls
\end{minipage} & \begin{minipage}[b]{\linewidth}\raggedleft
Provider tokens
\end{minipage} & \begin{minipage}[b]{\linewidth}\raggedleft
Episode time, s
\end{minipage} \\
\midrule\noalign{}
\endfirsthead
\multicolumn{6}{l}{\small\itshape Table A7 (continued)}\\
\toprule\noalign{}
\begin{minipage}[b]{\linewidth}\raggedright
Arm
\end{minipage} & \begin{minipage}[b]{\linewidth}\raggedleft
Valid deliveries
\end{minipage} & \begin{minipage}[b]{\linewidth}\raggedleft
Fresh BA attempts
\end{minipage} & \begin{minipage}[b]{\linewidth}\raggedleft
API calls
\end{minipage} & \begin{minipage}[b]{\linewidth}\raggedleft
Provider tokens
\end{minipage} & \begin{minipage}[b]{\linewidth}\raggedleft
Episode time, s
\end{minipage} \\
\midrule\noalign{}
\endhead
\bottomrule\noalign{}
\endlastfoot
R & 4/4 & 4 & 0 & N/A & 14.901 \\
L & 4/4 & 4 & 20 & 192,628 & 327.012 \\
G0 & 3/4 & 4 & 24 & 236,572 & 319.271 \\
GE & 4/4 & 4 & 25 & 360,230 & 390.101 \\
\end{longtable}

Shared preparation took a total of 430.755 s across four workers and is excluded from the episode times in Table A7. An additional failed host launch took 0.648 s and invoked neither a model API nor numerical candidate execution.

All fifteen valid deliveries selected BA. The remaining G0 response also named BA but included an extra field in its stop object and failed the strict schema; it remains a failed delivery, not evidence of a different geometric choice. GE also incorrectly treated the fixed candidate's not-run solver status as a failure of the BA usability requirement in one explanation, although that requirement applied only to BA. Thus successful schema compliance did not guarantee correct interpretation. All arms attempted the same number of fresh BA optimizations. The records document the executed feedback and state updates for each episode. The 69 calls are repeated decisions within four tasks, not independent experimental samples.

All 16 BA attempts reported usable solutions without convergence. In HKisland01-q75 and AMtown01-q25, training bounding-box diagonals grew from approximately 300 and 215 input units to \(5.34 \times 10^7\) and \(2.07 \times 10^8\), despite passing the pixel-based qualification. The units are not certified metric units. This discrepancy identifies training-geometry diagnosis as an additional prerequisite for dense reconstruction; the reported study retains its original binary candidate domain and qualification rule.

\subsection{Complete-chain mesh-parameter development check}\label{complete-chain-mesh-parameter-development-check}

A separate Dataset-2 check reran the complete fixed-camera numerical chain with mesh decimation 0.25, against the existing 0.5 product rescored under the same protocol. Face counts were 4,948,795 versus 9,802,046. Surface L1/RMSE were 0.232985/0.698582 m versus 0.231228/0.673999 m; best-90\% L1/RMSE were 0.125330/0.166690 m versus 0.125429/0.166851 m. Precision at 0.20 m was 71.7597\% versus 71.8267\%, and completeness was 69.863230\% versus 69.663557\%.

These are mixed single-product differences. Sparse-camera files were identical, while independently rerun dense products differed, preventing isolation of mesh decimation. An earlier scheduling interruption is accounted separately; the differently scheduled executions do not support a speed ratio.

\subsection{Reproduction source index}\label{reproduction-source-index}

The paths in Table A8 are relative to the research repository. Complete execution and session records remain in the local archive; the v0.1.0 archive scope is given in Appendix A.2. Candidate identities are determined from the program and score files.

\Needspace{7\baselineskip}
\textbf{Table A8. Selected local source locations.} This index consolidates proposal, camera, recommendation and reviewer records cited in Appendices A.3--A.11. It does not enumerate every artifact from the historical P0/P1 comparisons or the extended studies in Appendices A.12--A.14; the native-pilot P0 entries refer to contemporary controls.

\begin{longtable}[]{@{}
  >{\raggedright\arraybackslash}p{(\linewidth - 2\tabcolsep) * \real{0.5000}}
  >{\raggedright\arraybackslash}p{(\linewidth - 2\tabcolsep) * \real{0.5000}}@{}}
\toprule\noalign{}
\begin{minipage}[b]{\linewidth}\raggedright
Study or component
\end{minipage} & \begin{minipage}[b]{\linewidth}\raggedright
Source location and records
\end{minipage} \\
\midrule\noalign{}
\endfirsthead
\multicolumn{2}{l}{\small\itshape Table A8 (continued)}\\
\toprule\noalign{}
\begin{minipage}[b]{\linewidth}\raggedright
Study or component
\end{minipage} & \begin{minipage}[b]{\linewidth}\raggedright
Source location and records
\end{minipage} \\
\midrule\noalign{}
\endhead
\bottomrule\noalign{}
\endlastfoot
P2 proposal and execution & \path{out/geopilot-learning-20260923/proposal-format-replay/}: \texttt{request.json}, \texttt{response.json}, \texttt{program.json}, \texttt{proposal.json}, \texttt{context.json}, \texttt{candidate.json}; \path{p2-run/}, \path{p2-score-r2/}, \texttt{comparison-r2.json} under the same dated directory \\
Context and program validation & \path{experiments/geopilot_rsih_learning/continuation.py}; \path{code/geopilot_rsih/program.mjs} \\
Dataset-1 camera comparisons & \path{out/geopilot-research-20260925/}: \path{camera-six-metric-scores/}, \path{publisher-raster-full-sparse-run/}, \path{publisher-raster-mesh-r2-run/}, \path{publisher-raster-r2-scores/} \\
Publisher-camera implementation & \path{experiments/geopilot_rsih_learning/}: \texttt{publisher\_raster\_full\_sparse.py}, \texttt{publisher\_raster\_mesh\_r2.py}, \texttt{publisher\_raster\_score\_r2.py} \\
Native proposal pilot & \path{out/geopilot-research-20260925/native-post-prepare-pilot/}: \texttt{pack-a.json}, \texttt{pack-b.json}, \texttt{response-a.txt}, \texttt{response-b.txt}, \path{import-a-r2/program.json}, \path{import-b-r2/program.json} \\
Native-pilot execution status & Same pilot directory: \texttt{candidate-a-result-review.json}, \path{candidate-b-run/run.json}, \path{candidate-b-run/supervision.json}, \path{candidate-b-artifact-diagnostic/independent-result-review.json}, \path{p0-run/result.json}, \texttt{p0-serial-control-result-review.json} \\
Camera hypothesis and timeline & \path{out/geopilot-research-20260926/camera-boundary/decision-trace.md}; \texttt{preparation-diagnosis.md}, \texttt{shared-camera-ba-plan.md}, \texttt{camera-forward-plan.md} in the 25 September research directory \\
Camera diagnosis provenance & Camera-diagnosis session and continuation, locally archived; experimental sequence and source documents are described in Appendix A.6 \\
Dataset-2 recommendations & \path{out/geopilot-research-20260926/experience-ablation/}: \texttt{plan.md}, \texttt{plan-review.md}, \texttt{frozen.json}, \texttt{r01-packet.json}--\texttt{r09-packet.json}, original responses, \texttt{native-receipts.json}, \texttt{access-audit.json}, \texttt{results.json}, \texttt{response-review.json} \\
Preliminary reviewer study & \path{out/geopilot-jev-shadow-20260923-batch/}, \path{out/geopilot-jev-shadow-20260923-jev/}, \path{out/geopilot-jev-shadow-20260923-minimax/}; \path{experiments/geopilot_jev_shadow/replay.py} \\
Five reviewer batches & \path{out/geopilot-research-20260926/jev-validation/}: \texttt{text-live-v1}, \texttt{text-live-v2}, \texttt{text-live-evidence-v1}, \texttt{text-live-target-v1}, \texttt{text-live-evidence-target-v1} \\
Experience and diagnostic routing & \path{out/geopilot-research-20260928/decision-routing/}: \texttt{plan.md}, \texttt{results.json}, per-condition requests and responses, code snapshots and content review \\
End-to-end reviewer experiment & \path{out/geopilot-research-20260926/jev-validation/}: \texttt{system-pilot-protocol.md}, \path{system-pilot/ready-v4/}, per-run records, \texttt{summarize\_system.py}, \texttt{system-results.json}, \texttt{system-results.csv}, \texttt{system-tables.md} \\
\end{longtable}

\Needspace{18\baselineskip}
\textbf{Table A9. P2 proposal checksums.} The request uses schema \path{geopilot-offline-request/2}. Tool-knowledge and ordered-history checksums use the canonical JSON serialization described in Appendix A.3. The request checksum identifies the original file bytes; the instruction checksum identifies the system-instruction text encoded as UTF-8.

\begin{longtable}[]{@{}
  >{\raggedright\arraybackslash}p{(\linewidth - 2\tabcolsep) * \real{0.5000}}
  >{\raggedright\arraybackslash}p{(\linewidth - 2\tabcolsep) * \real{0.5000}}@{}}
\toprule\noalign{}
\begin{minipage}[b]{\linewidth}\raggedright
Component
\end{minipage} & \begin{minipage}[b]{\linewidth}\raggedright
SHA-256
\end{minipage} \\
\midrule\noalign{}
\endfirsthead
\multicolumn{2}{l}{\small\itshape Table A9 (continued)}\\
\toprule\noalign{}
\begin{minipage}[b]{\linewidth}\raggedright
Component
\end{minipage} & \begin{minipage}[b]{\linewidth}\raggedright
SHA-256
\end{minipage} \\
\midrule\noalign{}
\endhead
\bottomrule\noalign{}
\endlastfoot
Original request & \texttt{\seqsplit{097d483ba704e9306fefc7aa9f97529b700ac14830e979cc20352918586ed58f}} \\
UTF-8 system instruction & \texttt{\seqsplit{15731b5c144e9d7f86d3f262a6b8b27e1353b5f8aedb23704722f22b399dc712}} \\
Tool knowledge & \texttt{\seqsplit{14d3c6ab60f70d3b112443ca2cc89a933a7698eab15d79f5e1510811a4eb4e65}} \\
Ordered history & \texttt{\seqsplit{4e0fb641cd6ace0574fdd479ef5886b49f9ec303513159e2d0e151b44179404a}} \\
\end{longtable}

The UseGeo evaluation repository snapshot was \texttt{\seqsplit{aa6897530e1827d2527d09f32147edb1e44c8542}}. OpenMVS settings and version are given in Section~IV-C; the remaining numerical environment is specified in Appendix A.7. Individual binary identities are recorded with each experiment.

\section{Extended Results and Component Studies}\label{extended-results-and-component-studies}

The spatial analysis is presented in Section~V-C and Figure 7. It uses area samples from the fixed A and B meshes in their native coordinates, with seed 20260925 rather than the 20260916 seed used for Table~III. Semantic-review results are presented in Section~V-F and Figure S5, with complete batch counts and timing definitions in Appendix A.9.

\subsection{Complete restricted-program comparisons}\label{complete-restricted-program-comparisons}

Restricted-program proposals were translated into numerical calls and their meshes received external scores. P2 retained P0 as its parent and used P1's scored history as development feedback. P1 was not promoted. Table B1 gives its six-metric vector. The v0.1.0 research archive preserves the measurement file and the program-outcome record with its conditional interpretation. The separate camera intervention is reported in Section~V-A.

P1 tests whether the dense-point count is at least 5,000,000. The high-count branch meshes with decimation 1, hole closing 30, spurious removal 10 and smoothing 0, then refines at resolution level 2 with maximum face area 32. The low-count and unknown branches mesh with decimation 0.5, hole closing 20, spurious removal 5 and smoothing 0, without refinement. The recorded run took the high-count branch and delivered the refined mesh. Thus P1 changes conditional meshing and refinement, not only mesh size.

The subsequent candidate, P2, retained P0 as its parent and changed only the explicit mesh decimation parameter from 0.25 to 0.5. It added neither a diagnostic branch nor refinement. Its path executed preparation, densification, meshing, and stopping. The output contains 6,320,258 triangles, compared with 3,167,341 for P0. This confirms a change in delivered mesh size, not an improvement in geometric agreement. Because preparation and densification were rerun, a single program edit does not mean that the upstream numerical artifacts were held constant.

Tables B1 and B2 report all six measures for P1 and P2 against their common historical P0. P1 worsened all six measures; P2 increased mesh size with mixed geometric changes. The common P0 is one reconstruction reused for comparison, not an independent replicate in each table.

\Needspace{17\baselineskip}
\textbf{Table B1. Dataset-1 restricted-program comparison: P0 and P1.} One reconstruction per program under the common six-metric protocol; pp denotes percentage points.

\begin{longtable}[]{@{}
  >{\raggedright\arraybackslash}p{(\linewidth - 6\tabcolsep) * \real{0.4000}}
  >{\raggedleft\arraybackslash}p{(\linewidth - 6\tabcolsep) * \real{0.2000}}
  >{\raggedleft\arraybackslash}p{(\linewidth - 6\tabcolsep) * \real{0.2000}}
  >{\raggedleft\arraybackslash}p{(\linewidth - 6\tabcolsep) * \real{0.2000}}@{}}
\toprule\noalign{}
\begin{minipage}[b]{\linewidth}\raggedright
Metric
\end{minipage} & \begin{minipage}[b]{\linewidth}\raggedleft
Fixed P0
\end{minipage} & \begin{minipage}[b]{\linewidth}\raggedleft
Candidate P1
\end{minipage} & \begin{minipage}[b]{\linewidth}\raggedleft
P1 \(-\) P0
\end{minipage} \\
\midrule\noalign{}
\endfirsthead
\multicolumn{4}{l}{\small\itshape Table B1 (continued)}\\
\toprule\noalign{}
\begin{minipage}[b]{\linewidth}\raggedright
Metric
\end{minipage} & \begin{minipage}[b]{\linewidth}\raggedleft
Fixed P0
\end{minipage} & \begin{minipage}[b]{\linewidth}\raggedleft
Candidate P1
\end{minipage} & \begin{minipage}[b]{\linewidth}\raggedleft
P1 \(-\) P0
\end{minipage} \\
\midrule\noalign{}
\endhead
\bottomrule\noalign{}
\endlastfoot
Surface L1, m \(\downarrow\) & 1.988085 & 2.019519 & +0.031434 \\
Surface RMSE, m \(\downarrow\) & 2.392095 & 2.417309 & +0.025215 \\
Best-90\% L1, m \(\downarrow\) & 1.707988 & 1.741331 & +0.033343 \\
Best-90\% RMSE, m \(\downarrow\) & 2.017265 & 2.046680 & +0.029415 \\
Precision at 0.20 m, \% \(\uparrow\) & 6.406500 & 6.119500 & -0.287000 pp \\
Completeness at 0.20 m, \% \(\uparrow\) & 2.369538 & 2.244838 & -0.124700 pp \\
\end{longtable}

\Needspace{17\baselineskip}
\textbf{Table B2. Dataset-1 restricted-program comparison: P0 and P2.} One reconstruction per program under the common six-metric protocol; pp denotes percentage points.

\begin{longtable}[]{@{}
  >{\raggedright\arraybackslash}p{(\linewidth - 6\tabcolsep) * \real{0.4000}}
  >{\raggedleft\arraybackslash}p{(\linewidth - 6\tabcolsep) * \real{0.2000}}
  >{\raggedleft\arraybackslash}p{(\linewidth - 6\tabcolsep) * \real{0.2000}}
  >{\raggedleft\arraybackslash}p{(\linewidth - 6\tabcolsep) * \real{0.2000}}@{}}
\toprule\noalign{}
\begin{minipage}[b]{\linewidth}\raggedright
Metric
\end{minipage} & \begin{minipage}[b]{\linewidth}\raggedleft
Fixed P0
\end{minipage} & \begin{minipage}[b]{\linewidth}\raggedleft
Candidate P2
\end{minipage} & \begin{minipage}[b]{\linewidth}\raggedleft
P2 \(-\) P0
\end{minipage} \\
\midrule\noalign{}
\endfirsthead
\multicolumn{4}{l}{\small\itshape Table B2 (continued)}\\
\toprule\noalign{}
\begin{minipage}[b]{\linewidth}\raggedright
Metric
\end{minipage} & \begin{minipage}[b]{\linewidth}\raggedleft
Fixed P0
\end{minipage} & \begin{minipage}[b]{\linewidth}\raggedleft
Candidate P2
\end{minipage} & \begin{minipage}[b]{\linewidth}\raggedleft
P2 \(-\) P0
\end{minipage} \\
\midrule\noalign{}
\endhead
\bottomrule\noalign{}
\endlastfoot
Surface L1, m \(\downarrow\) & 1.988085 & 1.993562 & +0.005478 \\
Surface RMSE, m \(\downarrow\) & 2.392095 & 2.396214 & +0.004120 \\
Best-90\% L1, m \(\downarrow\) & 1.707988 & 1.714124 & +0.006136 \\
Best-90\% RMSE, m \(\downarrow\) & 2.017265 & 2.022759 & +0.005494 \\
Precision at 0.20 m, \% \(\uparrow\) & 6.406500 & 6.409800 & +0.003300 pp \\
Completeness at 0.20 m, \% \(\uparrow\) & 2.369538 & 2.370035 & +0.000497 pp \\
\end{longtable}

P2 increases the four distance errors and slightly increases both threshold measures. Its L1 difference is approximately +5.48 mm; precision and completeness differences are +0.003300 and +0.000497 percentage points. Differences use unrounded values before display rounding. With one reconstruction per program, a shared historical baseline and different host OS versions, repeatability and practical importance remain undetermined. P0 is retained.

The separate native proposal pilot has delivery observations but no complete comparative geometric result. Table A3 gives every attempt, including execution failures and the recovered artifact.

\subsection{Experience recommendation component}\label{experience-recommendation-component}

All three conditions selected F-U in each of their three calls and returned valid registered configurations (Section~IV-D; Table B3). Experience-conditioned responses cited the earlier U-raster comparison and distinguished it from M-raster intrinsic sharing. Direct responses reached the same choice from current input compatibility. Experience was reused in the explanations while the selected strategy remained unchanged. These nine proposals concern one task; they are not reconstruction replicates and have no associated Dataset-2 geometric scores. Complete responses and checks remain in the research record.

\Needspace{20\baselineskip}
\textbf{Table B3. Dataset-2 recommendation-component observations.} Each row contains three fresh native responses to one task. Registered-configuration and independent content checks assess the recommendation, not its geometric quality.

\begin{longtable}[]{@{}
  >{\raggedright\arraybackslash}p{(\linewidth - 10\tabcolsep) * \real{0.1429}}
  >{\raggedleft\arraybackslash}p{(\linewidth - 10\tabcolsep) * \real{0.1905}}
  >{\raggedright\arraybackslash}p{(\linewidth - 10\tabcolsep) * \real{0.1429}}
  >{\raggedleft\arraybackslash}p{(\linewidth - 10\tabcolsep) * \real{0.1905}}
  >{\raggedleft\arraybackslash}p{(\linewidth - 10\tabcolsep) * \real{0.1905}}
  >{\raggedright\arraybackslash}p{(\linewidth - 10\tabcolsep) * \real{0.1429}}@{}}
\toprule\noalign{}
\begin{minipage}[b]{\linewidth}\raggedright
Proposal context
\end{minipage} & \begin{minipage}[b]{\linewidth}\raggedleft
Responses
\end{minipage} & \begin{minipage}[b]{\linewidth}\raggedright
First choice
\end{minipage} & \begin{minipage}[b]{\linewidth}\raggedleft
Valid configurations
\end{minipage} & \begin{minipage}[b]{\linewidth}\raggedleft
Content violations
\end{minipage} & \begin{minipage}[b]{\linewidth}\raggedright
Geometry scored
\end{minipage} \\
\midrule\noalign{}
\endfirsthead
\multicolumn{6}{l}{\small\itshape Table B3 (continued)}\\
\toprule\noalign{}
\begin{minipage}[b]{\linewidth}\raggedright
Proposal context
\end{minipage} & \begin{minipage}[b]{\linewidth}\raggedleft
Responses
\end{minipage} & \begin{minipage}[b]{\linewidth}\raggedright
First choice
\end{minipage} & \begin{minipage}[b]{\linewidth}\raggedleft
Valid configurations
\end{minipage} & \begin{minipage}[b]{\linewidth}\raggedleft
Content violations
\end{minipage} & \begin{minipage}[b]{\linewidth}\raggedright
Geometry scored
\end{minipage} \\
\midrule\noalign{}
\endhead
\bottomrule\noalign{}
\endlastfoot
Direct model analysis & 3 & F-U: 3/3 & 3/3 valid & 0/3 & Not executed \\
GeoPilot procedure, no experience & 3 & F-U: 3/3 & 3/3 valid & 0/3 & Not executed \\
Same procedure with frozen experience & 3 & F-U: 3/3 & 3/3 valid & 0/3 & Not executed \\
\end{longtable}

Current-task compatibility already supplied a reason to choose F-U. Appendix A.8 identifies the frozen packets, responses and content checks.

\subsection{Complete-system review comparison}\label{complete-system-review-comparison}

Section~V-G of this paper and Figure S6 describe the three complete episodes, unchanged F-U actions and component times. Table B4 retains every geometric measure; Table A5 gives resource accounting.

\Needspace{17\baselineskip}
\textbf{Table B4. Complete Dataset-2 development outcomes under the frozen review protocol.} R0 uses no added semantic review; RF uses MiniMax review; RJ uses Jev with MiniMax fallback. Each condition independently generated one mesh with the same F-U branch and fixed numerical settings. All three delivered valid full scores and complete recorded image-processing coverage. Lower distance errors and higher precision/completeness are favorable. Each condition was run once; the differences are descriptive.

\begin{longtable}[]{@{}lrrr@{}}
\toprule\noalign{}
Measure & R0 & RF & RJ \\
\midrule\noalign{}
\endfirsthead
\multicolumn{4}{l}{\small\itshape Table B4 (continued)}\\
\toprule\noalign{}
Measure & R0 & RF & RJ \\
\midrule\noalign{}
\endhead
\bottomrule\noalign{}
\endlastfoot
Surface L1, m & 0.232631 & 0.231508 & 0.231882 \\
Surface RMSE, m & 0.693818 & 0.687521 & 0.691681 \\
Best-90\% L1, m & 0.125353 & 0.125092 & 0.125199 \\
Best-90\% RMSE, m & 0.166716 & 0.166361 & 0.166525 \\
Precision at 0.20 m, \% & 71.740000 & 71.864200 & 71.869800 \\
Completeness at 0.20 m, \% & 69.807339 & 69.868023 & 69.670118 \\
\end{longtable}

\clearpage
\section{Additional Workflow and Component Figures}\label{appendix-c.-additional-workflow-and-component-figures}

\begin{figure}[!htbp]\renewcommand{\thefigure}{S\arabic{figure}}
\centering
\pandocbounded{\includegraphics[width=\linewidth,height=0.68\textheight,keepaspectratio]{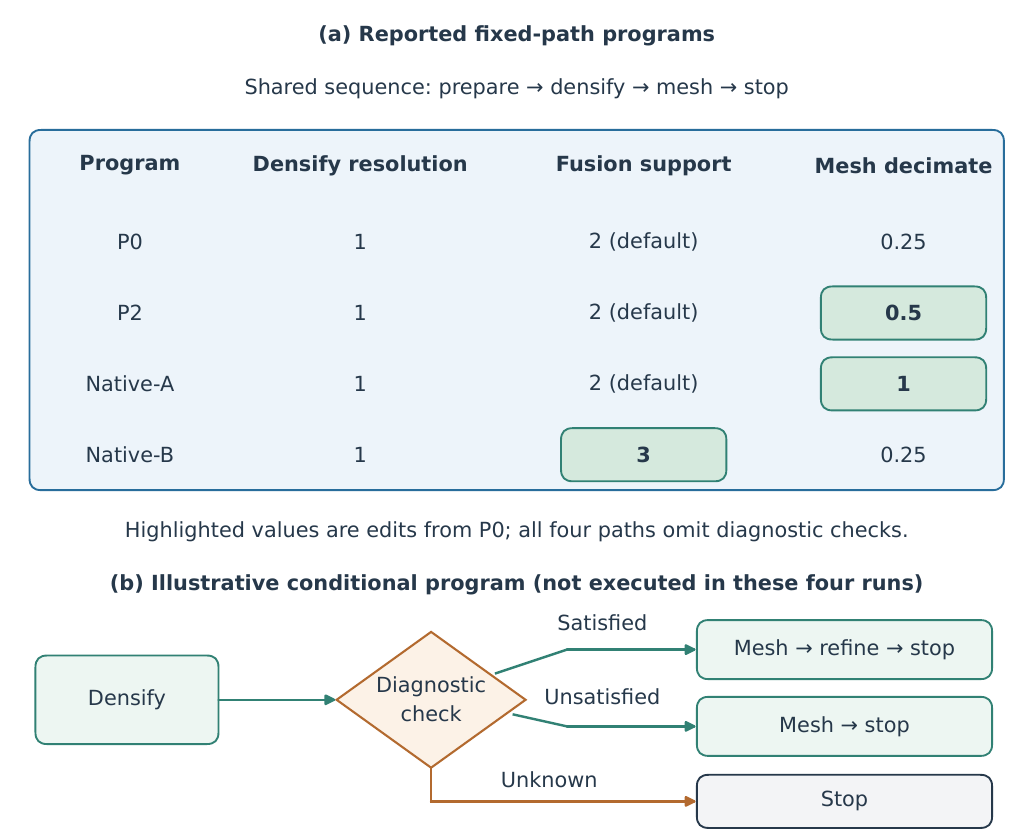}}
\caption[]{The executed P0, P2, Native-A and Native-B programs use fixed paths, with parameter changes highlighted. The separate unexecuted diagnostic branch illustrates conditional graph syntax.}
\end{figure}

\begin{figure}[!htbp]\renewcommand{\thefigure}{S\arabic{figure}}
\centering
\pandocbounded{\includegraphics[width=\linewidth,height=0.68\textheight,keepaspectratio]{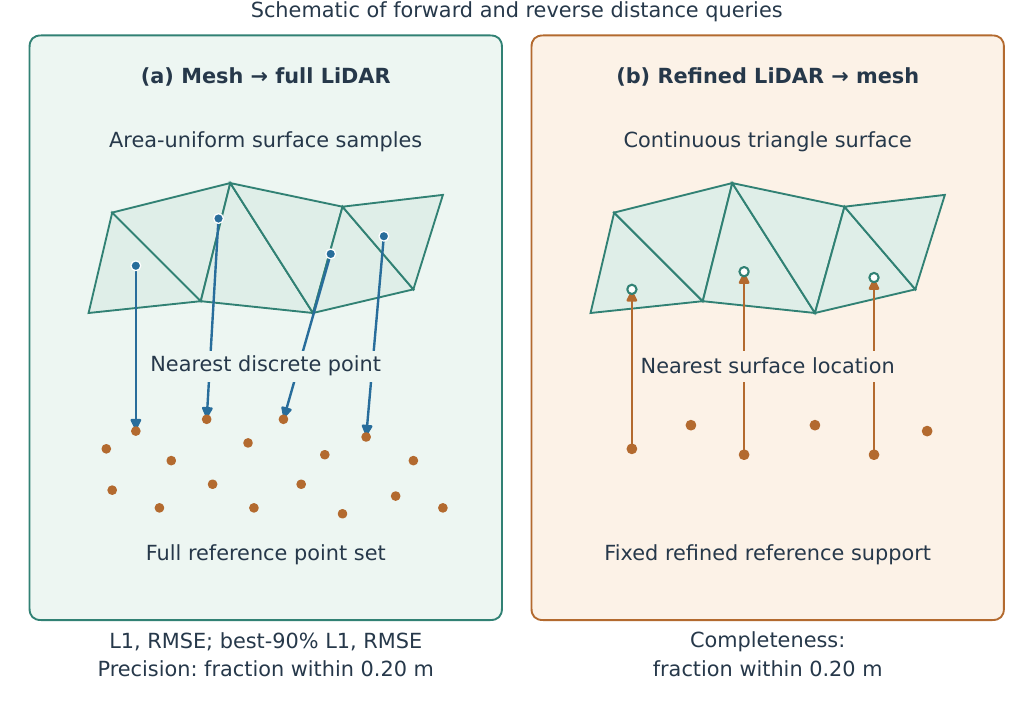}}
\caption[]{Forward distances query the full LiDAR reference from area-uniform mesh samples; reverse completeness queries the triangle surface from the fixed refined reference. Precision and completeness use 0.20 m. Refined support follows the publisher-MVS-derived 5 m XY rule, rather than a verified visibility mask.}
\end{figure}

\begin{figure}[!htbp]\renewcommand{\thefigure}{S\arabic{figure}}
\centering
\pandocbounded{\includegraphics[width=\linewidth,height=0.68\textheight,keepaspectratio]{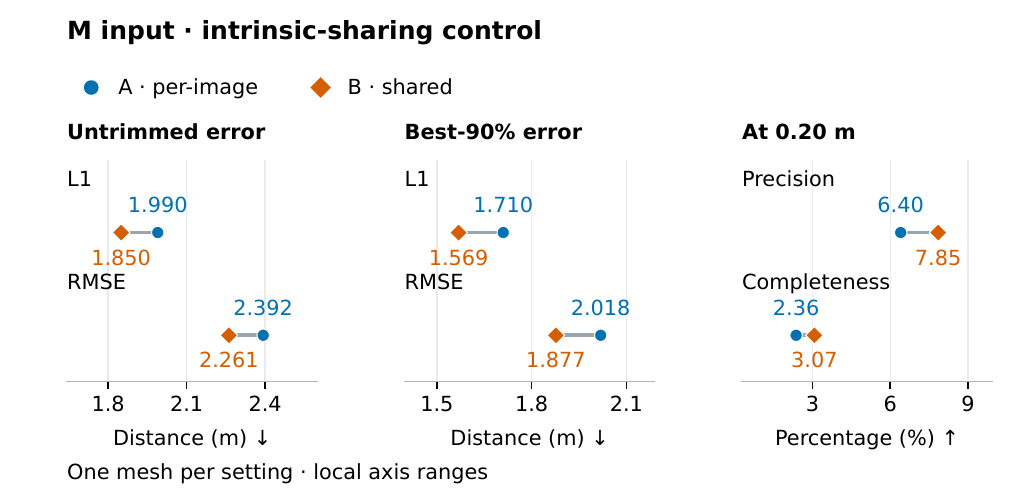}}
\caption[]{Shared intrinsics improve all six native-frame measures on Dataset-1 M. Lines connect A (per-image) and B (shared); distance units are meters and threshold measures are percentages at 0.20 m. Axes use local ranges. Each configuration has one reconstruction, so reconstruction-repeat intervals are unavailable. Figure 9 shows the reference-assisted reversal of the mean-distance ordering.}
\end{figure}

\begin{figure}[!htbp]\renewcommand{\thefigure}{S\arabic{figure}}
\centering
\pandocbounded{\includegraphics[width=\linewidth,height=0.68\textheight,keepaspectratio]{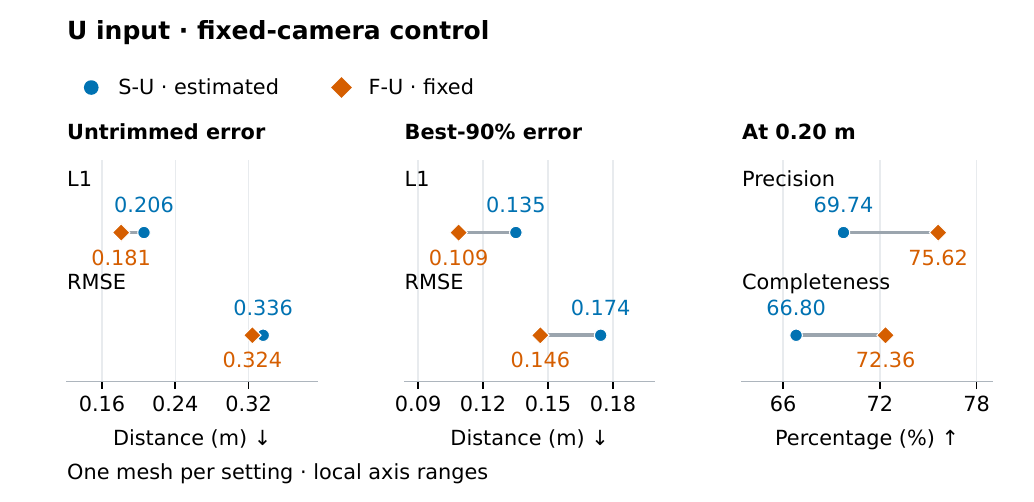}}
\caption[]{Fixed published cameras improve all six measures relative to estimated cameras on identical Dataset-1 U pixels. Lines connect S-U and F-U, with distances in meters and threshold measures in percent at 0.20 m. Axes use local ranges. Each configuration has one reconstruction; F-U retains the published cameras' upstream adjustment dependence. M and U absolute scores concern different inputs.}
\end{figure}

\begin{figure}[!htbp]\renewcommand{\thefigure}{S\arabic{figure}}
\centering
\pandocbounded{\includegraphics[width=\linewidth,height=0.68\textheight,keepaspectratio]{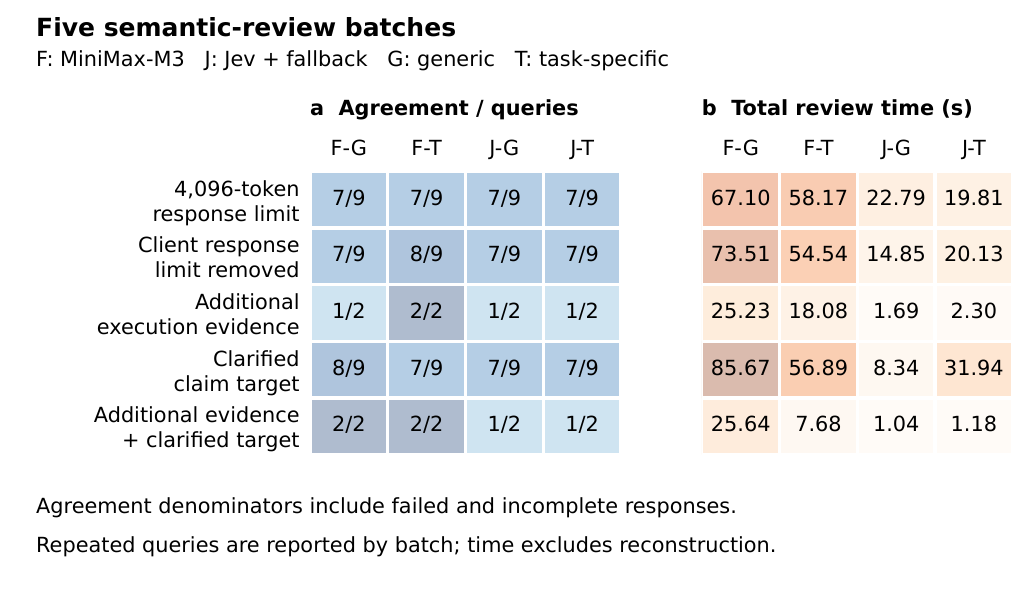}}
\caption[]{Agreement and summed review time in five semantic-review batches. F is MiniMax-M3, J is Jev with MiniMax fallback, and G/T denote generic/task-specific rubrics. Response limits refer to the MiniMax client. Denominators include failed and incomplete responses: nine queries in batches 1, 2 and 4, and two selected diagnostic queries in batches 3 and 5. Blue represents agreement fractions (0--1); orange represents seconds (0--90). Time includes preparation, calls, parsing, fallback and saving, and excludes shared loading, rubric/evidence construction, annotation and reconstruction/scoring. Batches are reported separately because repeated questions and selected follow-ups are correlated.}
\end{figure}

\begin{figure}[!htbp]\renewcommand{\thefigure}{S\arabic{figure}}
\centering
\pandocbounded{\includegraphics[width=\linewidth,height=0.68\textheight,keepaspectratio]{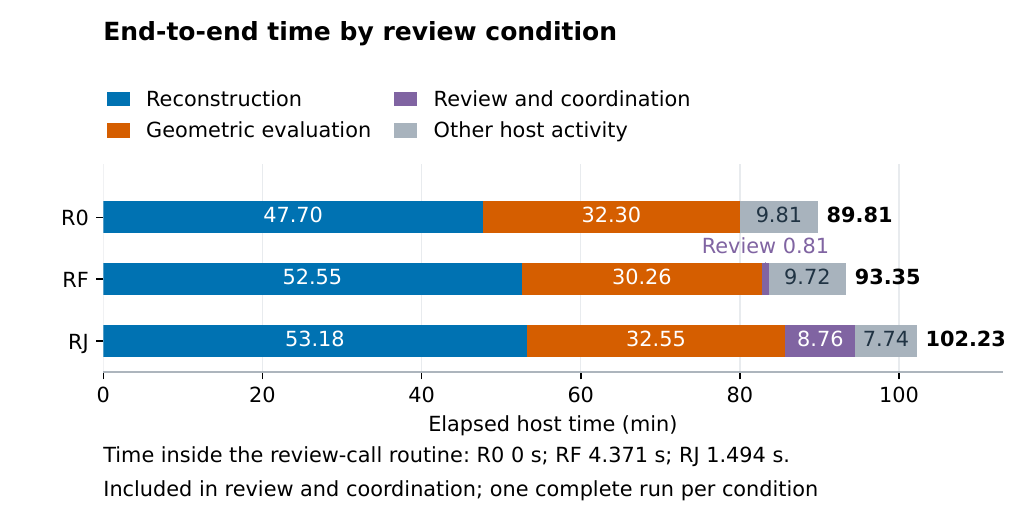}}
\caption[]{Reconstruction and scoring dominate the complete elapsed times of 89.81, 93.35 and 102.23 min for R0, RF and RJ. Each bar represents one independently reconstructed and fully scored Dataset-2 F-U mesh. Stacks separate numerical reconstruction, full scoring, review and coordination intervals, and other host activity. Time inside the review-call routine is included in review and coordination. These single runs with uncontrolled provider, host and reconstruction variability do not establish a stable speed difference.}
\end{figure}

\clearpage
\bibliographystyle{IEEEtran}
\bibliography{references}
\end{document}